\documentclass{article}
\usepackage{ijcai25}

\usepackage{times}
\usepackage{soul}
\usepackage{url}
\usepackage[hidelinks]{hyperref}
\usepackage[utf8]{inputenc}
\usepackage[small]{caption}
\usepackage{graphicx}
\usepackage{amsmath}
\usepackage{amsthm}
\usepackage{booktabs}
\usepackage{algorithm}
\usepackage{algorithmic}
\usepackage[switch]{lineno}

\usepackage{subcaption}  
\usepackage{makecell}
\usepackage {amsfonts,amssymb}
\usepackage{booktabs}
\usepackage{multirow}
\usepackage{xcolor} 
\usepackage{makecell} 
\usepackage{float}
\usepackage{mathtools}
\usepackage{tikz}
\newcommand*{\circled}[1]{\lower.7ex\hbox{\tikz\draw (0pt, 0pt)%
    circle (.5em) node {\makebox[0em][c]{\small #1}};}}

\newtheorem{theorem}{Theorem}

\newtheorem{proposition}[theorem]{Proposition}  
\newtheorem{definition}{Definition}

\newtheorem{remark}{Remark}
\usepackage{bbm}

\title{Instructing Text-to-Image Diffusion Models via Classifier-Guided Semantic Optimization}

\author{
Yuanyuan Chang$^1$\and
Yinghua Yao$^{2,3}$\thanks{Corresponding author}\and
Tao Qin$^1$\and
Mengmeng Wang$^{4,5}$\and
Ivor Tsang$^{2,3}$\And
Guang Dai$^5$
\\
\affiliations
$^1$MOE Key Laboratory for Intelligent Networks and Network Security, Xi’an Jiaotong University\\
$^2$Center for Frontier AI Research, Agency for Science, Technology and Research, Singapore\\
$^3$Institute of High Performance Computing, Agency for Science, Technology and Research, Singapore\\
$^4$Zhejiang University of Technology
$^5$SGIT AI Lab, State Grid Corporation of China
\emails
1371306634@stu.xjtu.edu.cn,
eva.yh.yao@gmail.com,
qin.tao@mail.xjtu.edu.cn,
mengmengwang@zju.edu.cn,
 \{ivor.tsang, guang.gdai\}@gmail.com
}

\begin{document}

\maketitle

\begin{abstract}
Text-to-image diffusion models have emerged as powerful tools for high-quality image generation and editing. Many existing approaches rely on text prompts as editing guidance. However, these methods are constrained by the need for manual prompt crafting, which can be time-consuming, introduce irrelevant details, and significantly limit editing performance. In this work, we propose optimizing semantic embeddings guided by attribute classifiers to steer text-to-image models toward desired edits, without relying on text prompts or requiring any training or fine-tuning of the diffusion model. We utilize classifiers to learn precise semantic embeddings at the dataset level. The learned embeddings are theoretically justified as the optimal representation of attribute semantics, enabling disentangled and accurate edits. Experiments further demonstrate that our method achieves high levels of disentanglement and strong generalization across different domains of data. Code is available at \href{https://github.com/Chang-yuanyuan/CASO}{\textcolor{pink}{\textit{https://github.com/Chang-yuanyuan/CASO}}}.
\end{abstract}

\section{Introduction}
Faithful image reconstruction is a fundamental prerequisite for image editing tasks. Recently, diffusion-based \cite{DDPM,DDIM,LDM,song2021scorebased} generative models have demonstrated significant advantages over GANs \cite{gan} in the field of image generation \cite{clg}. These approaches have emerged as powerful tools for image editing due to their generative modeling capabilities. However, exploring the decoupled semantic subspace of diffusion models remains a critical yet challenging area of research.

Most of the initial works on diffusion-based image editing require training or fine-tuning the model and focus on a single domain of data \cite{diffusionautoencoders,HDAE}. The rapid rise in popularity of text-to-image diffusion models, such as Stable Diffusion (SD) \cite{LDM}, DeepFloyd IF \cite{DeepFloyd}, and Latent Consistency Models \cite{consistency},  has further inspired researchers to achieve decoupled image editing based on text or other conditions (e.g., semantic segmentation) without requiring additional training or fine-tuning \cite{SEGA,OIG}. Unlike small, single-domain diffusion models, these advanced models enable cross-domain generation, broadening their applicability across diverse image editing scenarios, have become a hot spot of current research.
\begin{figure}[tb]
    \centering
    \includegraphics[width=\columnwidth]{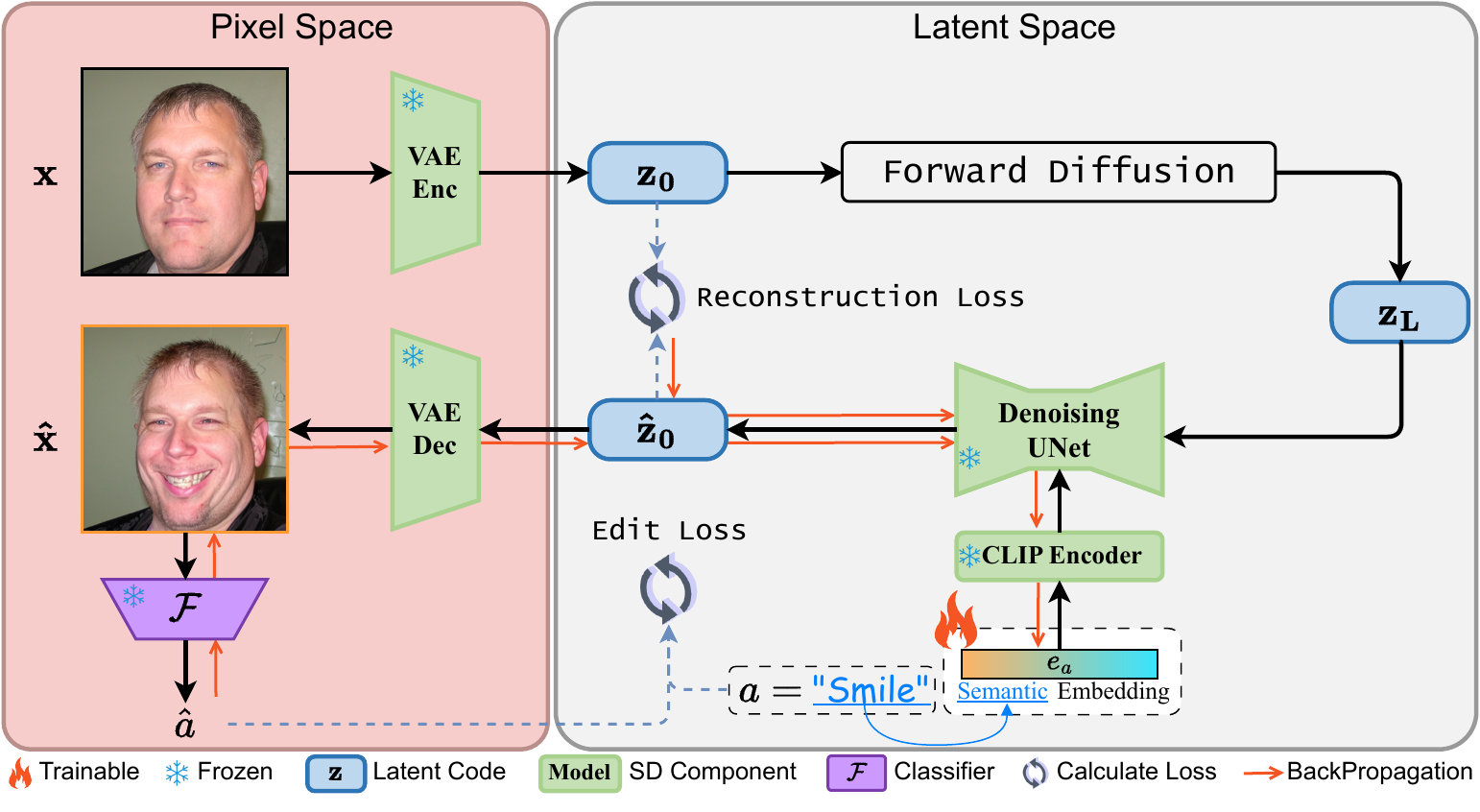} 
    \caption{\textbf{C}l\textbf{A}ssifier-guide \textbf{S}emantic \textbf{O}ptimization (CASO). The trainable continuous semantic embedding for the target attribute $a$, guides Stable Diffusion for desired edits.}
    \label{framework}
\end{figure}
\begin{figure*}[tb]
    \centering
    \includegraphics[width=0.9\textwidth]{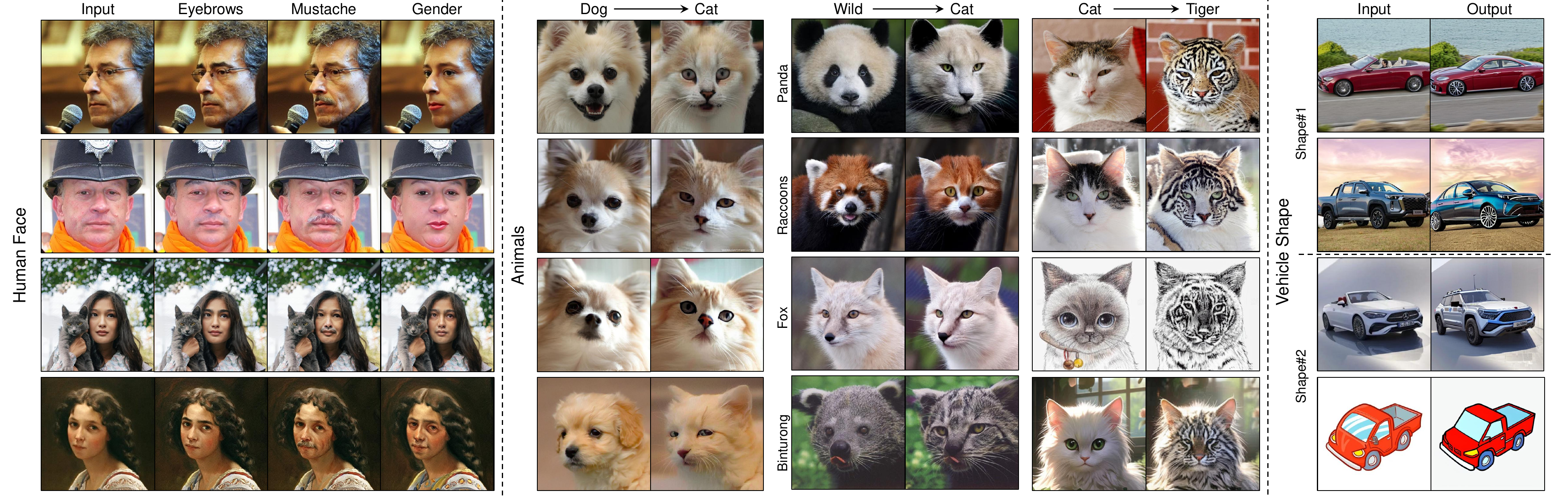} 
    \caption{\textbf{CASO Edit Result.} Our method generalizes well to data with different styles.}
    \label{edit}
\end{figure*}
However, existing solutions are mainly based on text prompts \cite{SEGA,cyclediffusion,OIG}, which present several challenges. Creating appropriate text prompts can be inherently difficult, and prompts with identical semantics often yield significantly divergent outcomes. Furthermore, achieving disentangled editing through text prompts alone is highly challenging, as they frequently affect unrelated regions of the image. Recent works that eschew text prompts proposed unsupervised ways to discover editing directions in the latent space of diffusion models \cite{loco,dalva2024noiseclr} also exhibit limitations, particularly in terms of generalization. For instance, while these methods explore editable directions using real face data, their applicability often diminishes when extended to other types of images, such as artistic or anime-style faces. Moreover, they can only obtain a limited number of subtle editing directions, often lack the ability to specify attributes, and rely on the clear structure or inherent variance within the dataset itself.

In this paper, we propose optimizing continuous semantic embeddings through attribute classifier gradients to guide text-to-image models for targeted editing. Our framework is shown in Fig.~\ref{framework}. It can identify disentangled and generalizable editing directions. 
Our contributions are as follows.
\begin{itemize}
\item We propose \textbf{C}l\textbf{A}ssifier-Guided \textbf{S}emantic \textbf{O}ptimization (CASO), a lightweight and efficient framework for image editing, which simply tunes trainable semantic embeddings and does not require training or fine-tuning of diffusion models.
\item We theoretically and experimentally demonstrate that CASO learns semantic embeddings containing attribute class mean information, thereby enabling precise attribute editing.
\item Experimental results show that our editing directions are highly disentangled and exhibit strong generalization, outperforming various existing approaches.
\end{itemize}

\section{Related Work}
Deep generative models have achieved great success in image editing.  Various types of generative models, such as VAEs \cite{vae}, GANs \cite{gan}, flow-based models \cite{nice}, and diffusion models \cite{DDPM,DDIM}, have been adopted as editing frameworks.  Among these, the recent state-of-the-art editing models are based on diffusion models.  In the following, we will focus on reviewing work on diffusion based image editing.
 \label{Diffusion based image edit}
 
 Diffusion models perform iterative denoising based on a sequence of latent variables, which lack explicit semantic representations like those in GANs. So identifying the disentangled semantic space in diffusion models remains a significant challenge. Some image editing methods based on diffusion models require retraining or fine-tuning a diffusion model \cite{diffusionclip,stylediffusion,diffusionautoencoders,unitune}. For example, DiffAE \cite{diffusionautoencoders} and HDAE \cite{HDAE} use diffusion model as the representation encoders and enable image editing by modifying representations. These methods are generally only suitable for single-domain (such as human face) diffusion models. DiffusionCLIP \cite{diffusionclip} fine-tunes the diffusion model with the help of the CLIP \cite{CLIP}. UniTune \cite{unitune} fine-tunes the diffusion model on a single base image, encouraging the model to generate images similar to the base image. However, training or fine-tuning the diffusion model can be somewhat expensive.

 Recent advances \cite{object,SEGA,plug,cyclediffusion,p2p,nulltext} avoid training or fine-tuning diffusion models.
 Some methods (SEGA \cite{SEGA} and OIG \cite{OIG})  use text prompts to guide editing. OIG uses CLIP text encoder \cite{CLIP} to ensure semantic correlation and extracts the feature map of UNet to ensure structural similarity. SEGA generates editing noise from text prompts and then extracts a small portion of this noise (by default, the top 5\%, based on empirical observations) for editing. Unfortunately, the 5\% threshold can only be set empirically, which is an unstable way to edit and can cause image quality loss. 
 Additionally, some work optimize text-based embeddings to faithfully reconstruct the input image, so as to achieve disentangled edits (the editing does not change the irrelevant features) with classifier-free guidance. 
Null-Text Inverse \cite{nulltext} fine-tunes the null-text embedding to align the sampling and inversion trajectories, enabling accurate image reconstruction, but it relies on a user-provided source prompt. Prompt Tuning Inversion \cite{prompt_tuning} introduces a faster approach by encoding image information into a learnable text embedding and performing editing through linear interpolation with the target embedding. However, these methods prioritize reconstruction quality, whereas our goal is to obtain an optimal semantic embedding for effective image editing.

 \begin{figure*}[htb]
    \centering
    \includegraphics[width=0.8\textwidth]{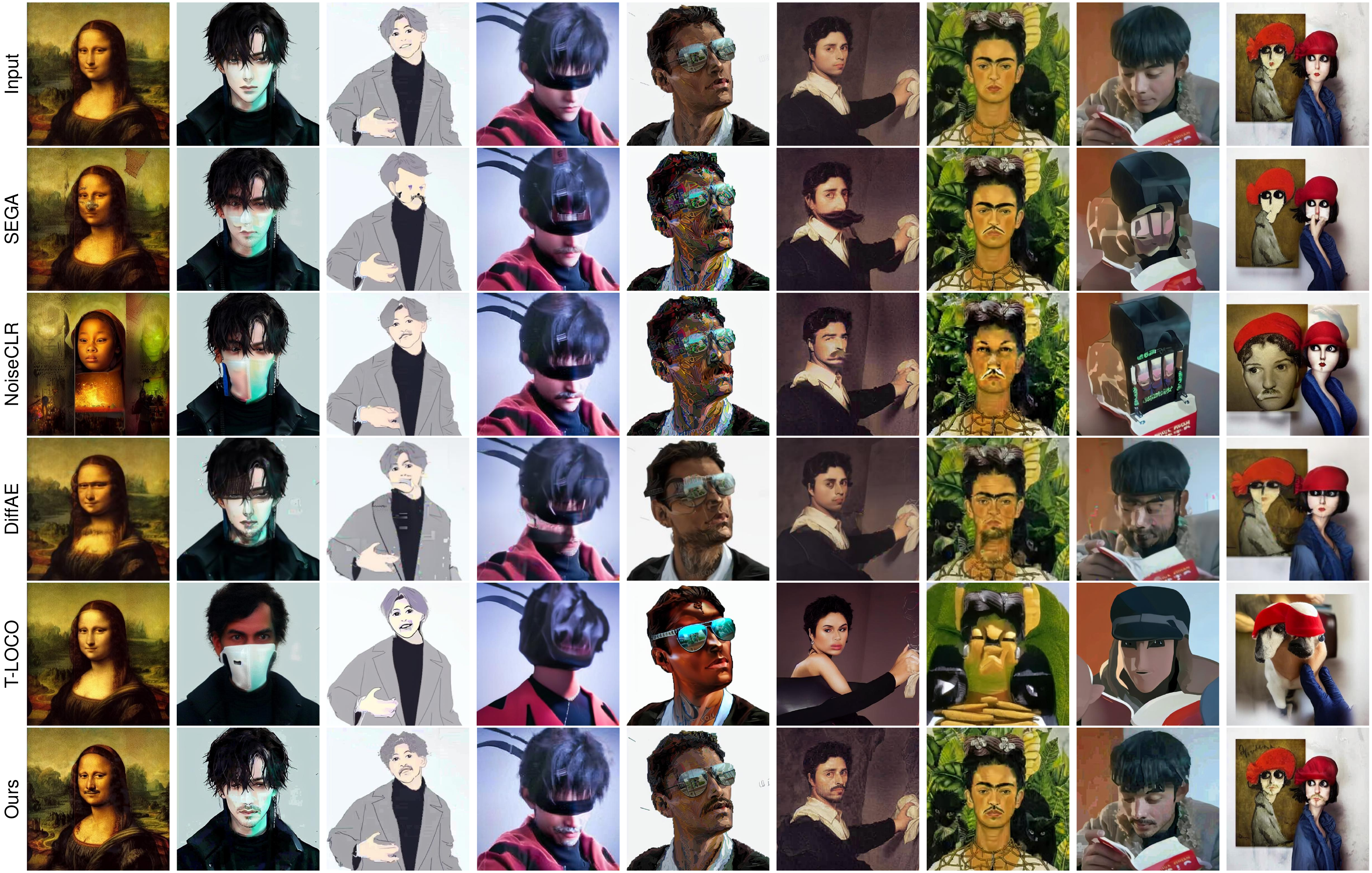} 
    \caption{\textbf{Comparison of different methods for attribute ``Mustache".} Our method shows the best generalization because it captures the exact semantics at the dataset level.}
    \label{compare}
\end{figure*}
 NoiseCLR \cite{dalva2024noiseclr} and LOCO \cite{loco} propose unsupervised methods to discover interpretable directions in pre-trained diffusion models, but they suffer from several shortcomings. The unsupervised learned direction is uncontrollable and may not converge to user-desired direction. And no more than 10\% of the directions learned by NoiseCLR are semantic meaningful. LOCO is only suitable for local editing, and some major structural changes such as cat$\rightarrow$dog are not feasible.

\section{Method}

 In this section, we describe our proposed method. First, we briefly discuss the background of latent diffusion models, and then introduce our method.

\subsection{Latent Diffusion and Classifier-free Guidance}
Diffusion models are generative models that produce data samples by fitting the data distribution through an iterative denoising process~\cite{DDPM}. Stable Diffusion (SD)~\cite{LDM} is a type of Latent Diffusion Model (LDM) that operates in the latent space of image data, where the conversion between latent code~$z$ and raw image data~$x$ is performed by the VAE encoder $\mathcal{E}$ and decoder $\mathcal{D}$. The training loss is defined as:
\begin{equation}
\mathcal{L}_{\text{LDM}}= \mathbb{E}_{z, \epsilon \sim \mathcal{N}(0,1), t} \left[ \left\| \epsilon - \epsilon_{\theta}(z_t, t) \right\|_2^2 \right],
\end{equation}
where $t$ is uniformly sampled from $\{1,\ldots,T \}$ and $z_t$ is a noisy version of ~$z_0=\mathcal{E}(x)$.

Classifier-free guidance \cite{cfg} trains a conditional diffusion model. Since the condition is left blank with a certain probability during training, it also supports unconditional generation.
The final predicted noise with classifier-free guidance in SD is defined as: 
\begin{equation}
\label{cfg}
\tilde{\epsilon}_{\theta}(z_t,c)=\epsilon_\theta(z_t, \phi) + \lambda (\epsilon_\theta(z_t,c) - \epsilon_\theta(z_t,\phi)),
\end{equation}
where $c$ is the condition, $\phi$ is null text and $\lambda$ is guidance scale.
For simplicity, we use $\epsilon_{\theta}(z_t)$ instead of $\epsilon_{\theta}(z_t, t)$ to represent the predicted noise for timestep $t$, as $t$ is implicitly denoted with variable $z_t$.

\subsection{Classifier-guided Semantic Optimization}
We can set $c$ as a suitable text to guide the editing. However, as shown in Appendix.B, even when sampling with the same random seed, images generated from different text prompts with the same semantic meaning exhibit significant differences (the first four images). Furthermore, adding additional descriptions to the text also leads to substantial changes in the overall appearance of the generated images (the first and last images). Text guidance struggles to achieve disentangled edits, namely, preserving the attribute excluding details. To achieve precise image editing, we propose a lightweight approach to obtain semantic embeddings as conditional guidance based on a ready-to-use attribute classifier.

Suppose the target editing attribute has $K$ classes, i.e., $a \in \{1, \ldots, K\}$. Define the ready-to-use attribute classifier as~$\mathcal{F}$ and the images generated by the Stable Diffusion editing process as~$\hat{x}=\mathcal{G}(x, e_{a})$, where $x$ is the input image and $e_a$ is the embedding for each class~$a$ of the attribute. We define an editing loss to optimize~$\{e_a\}_{a=1}^{K}$ as follows:
\begin{equation}
\label{cls_loss}
    \mathop{\min}_{\{e_a\}_{a=1}^K} \mathcal{L}_{\text{edit}}=\mathbb{E}_{x, a}\left[\ell_{\text{c}}\left(\mathcal{F}(\mathcal{G}(x, e_a)), a\right)\right]
\end{equation}
where $\ell_{\text{c}}$ is the classification loss that enforces generated images conditioned on $e_a$ to be classified as the corresponding attribute class~$a$.

The attribute embedding~$\{e_a\}_{a=1}^K$ obtained above contains the attribute class information at the dataset level, enabling image editing with precise semantics. We provide a theoretical justification in the following.

Denoting all layers of the classifier except the last as $\mathcal{F}_{\scriptscriptstyle -1}$, and the last layer with weight $w_a$ and bias $b_a$ for each attribute class $a$, the classifier for predicting a class label can be expressed as $\arg\max_{a'} \left(\langle w_{a'}, \mathcal{F}_{\scriptscriptstyle -1}(x) \rangle + b_{a'}\right)$.

\begin{definition}[Globally-centered attribute class mean~$\mu_a$] \label{definition}
We define the globally-centered attribute class mean~$\mu_a$ as:
\begin{equation*}
\mu_{a} = \mathbb{E}_{i}[h_{i, a}]-\mathbb{E}_{i, a}[h_{i, a}], 
\end{equation*}\label{w_a}
where $a=1, 2, \ldots, K$ and $h_{i, a}=\mathcal{F}_{\scriptscriptstyle -1}(x_{i, a})$ is the last-layer feature of the $i$-th sample in attribute class~$a$ for training the classifier.
\end{definition}

\begin{theorem}\cite{neuralcollapse} \label{Theorem 1}
For a sufficiently large classifier network, the last layer of the classifier~$w_a$ will converge to the globally-centered class mean~$\mu_a$, namely,
\begin{equation}
\label{our_nc2}
    \frac{w_{a}}{\left\|w_{a}\right\|_{2}} - \frac{\mu_{a}}{\left\|\mu_{a}\right\|_{2}} \rightarrow 0,
\end{equation}
where $a=1, 2, \ldots, K$. 
\hfill\rule{2mm}{2mm}
\end{theorem}

Theorem 1 describes the neural collapse phenomenon of training a deep classifier network~\cite{neuralcollapse,papyan2020prevalence,yang2022inducing}. We apply it to demonstrate that our classifier-guided semantic embedding encodes the attribute class mean. 

\begin{proposition} \label{Prop_2}
Give a sufficiently large classifier network~$\mathcal{F}$ which is well-trained on the training data. When optimizing $\{e_a\}_{a=1}^K$ using Eq.~\eqref{cls_loss} until converge while fixing $\mathcal{F}$, the neural collapse phenomenon still happens, namely, 
\begin{equation}\label{w_a_}
    \frac{w_{a}}{\left\|w_{a}\right\|_{2}}-\frac{\mu^{\prime}_{a}(e_a)}{\left\|\mu^{\prime}_{a}(e_a)\right\|_{2}}\rightarrow 0,
\end{equation}
where $a=1, 2, \ldots, K$. In particular,
\begin{equation*}
\begin{aligned}
    \mu^{\prime}_{a}(e_a) = \mathbb{E}_{i}\left[\mathcal{F}_{\scriptscriptstyle -1}(\mathcal{G}(x_{i}, e_a))\right] - \mathbb{E}_{i, a}\left[\mathcal{F}_{\scriptscriptstyle -1}(\mathcal{G}(x_{i}, e_a))\right],
\end{aligned}
\end{equation*}
which denotes the globally-centered attribute class mean of generated images.
\hfill\rule{2mm}{2mm}
\end{proposition}
The proof is similar to Theorem~\ref{Theorem 1} given the same classifier. We give a detailed proof in the Appendix. 

With Eq.~\eqref{our_nc2} and Eq.~\eqref{w_a_}, we can easily obtain $\frac{\mu^{\prime}_{a}(e_a)}{\left\|\mu^{\prime}_{a}(e_a)\right\|_{2}} \rightarrow \frac{\mu_{a}}{\left\|\mu_{a}\right\|_{2}}$. This means the optimal attribute embedding~$e_a$ in Eq.~\eqref{cls_loss} is determined by the attribute class mean~$\mu_{a}$. We also verify this in our empirical study.

\begin{table}
    \centering
    \begin{tabular}{lcccc}
    \toprule
    Method & Old & Mustache & Gender & Lipstick \\
    \midrule
    Comp-Diff & 0.19 & 0.40 & 0.42 & 0.38 \\
    DiffAE & 0.20 & 0.19 & 0.24 & 0.23  \\
    Cycle-Diff & \textbf{0.10} & 0.21 & 0.23 & 0.41 \\
    SEGA & \underline{0.11} & 0.23 & 0.27 & 0.42 \\
    NoiseCLR & 0.17 & \underline{0.17} & \textbf{0.20} & \underline{0.19} \\
    T-LOCO & 0.26 & 0.24 & 0.28 & 0.24 \\ 
    Ours & 0.16 & \textbf{0.13} & \textbf{0.20} & \textbf{0.18} \\
        \bottomrule
    \end{tabular}
    \caption{Comparison of various methods with LPIPS($\downarrow$) for human face edit.}
    \label{tab:comparison}
\end{table}

\subsubsection{Image Editing by Diffusion Models}
The typical image editing pipeline using LDMs incorporates conditioning during the denoising process, following these steps \cite{nulltext}:  \\
({\romannumeral1}) The input image $x$ is encoded into its latent code $z_0 = \mathcal{E}(x)$ using the VAE encoder $\mathcal{E}$.  \\
({\romannumeral2}) The clean latent code $z_0$ is converted to its noisy counterpart $z_L$ through DDIM inversion (Eq.~\eqref{nulltextinversion}), a deterministic mapping process.  \\
\begin{equation} \label{nulltextinversion}
    z_{t+1} = \sqrt{\frac{\alpha_{t+1}}{\alpha_t}} z_t + \left( \sqrt{\frac{1}{\alpha_{t+1}} - 1} - \sqrt{\frac{1}{\alpha_t} - 1} \right) \cdot \epsilon_\theta(z_t, \phi),
\end{equation}
where $\alpha_t$ is the time dependent scale.\\
({\romannumeral3}) Classifier-free guidance (Eq.~\eqref{cfg}) is applied during the denoising process using the condition $e_a$:  
\begin{align} \label{ddim sample}
    \hat{z}_{t-1} =& \sqrt{\alpha_{t-1}}\left( \frac{\hat{z}_t-\sqrt{1-\alpha_t}\tilde{\epsilon}_{\theta}(\hat{z}_t, e_a)}{\sqrt{\alpha_t}} \right)  \nonumber\\ 
    &+ \sqrt{1-\alpha_{t-1}} \tilde{\epsilon}_{\theta}(\hat{z}_t, e_a) ,
\end{align}
where $t=L, L-1, \ldots, 1$ and $\Tilde{\epsilon}_{\theta}(z_t,e_a)$ is calculated according to Eq.~\eqref{cfg}. 
We initialize $\hat{z}_L = z_L$. \\
({\romannumeral4}) The edited image is obtained by decoding the denoised latent code~$\mathcal{D}(\hat{z}_0)$.

\begin{algorithm}[tb] 
    \caption{Training algorithm}
    \label{algorithm}
    \textbf{Input}: Training data $\mathcal{X}=\{x^i\}_{i=1}^N$,  batch size~$B$, denoising UNet $\epsilon_\theta$, VAE encoder $\mathcal{E}$, VAE decoder $\mathcal{D}$, classifier $\mathcal{F}$. \\
    \textbf{Output}: Semantic embedding~\(e_a\)
    \begin{algorithmic}[1] 
    \REPEAT
        \STATE Sample a batch of images $\{x^i\}_{i=1}^{B}$.
        \STATE Get latent codes $\{z_0^i=\mathcal{E}(x^i)\}_{i=1}^{B}$.
        \STATE Following Eq.~\eqref{addnoise}, generate noise latent codes $\{z_{L}^i\}_{i=1}^B$.
        \STATE Following Eq.~\eqref{predx0}, generate edited images~$\{\hat{x}^i\}_{i=1}^B$.
        \STATE Train the condition embedding $e_a$ according to Eq.~\eqref{final_loss}.
    \UNTIL {converged} 
    \end{algorithmic}
\end{algorithm}
For training, the edited image~$\mathcal{D}(\hat{z}_0)$ is used as $\mathcal{G}(x, e_a)$ in Eq.~\eqref{cls_loss} to update the embedding.
However, steps ({\romannumeral2}) and ({\romannumeral3}) require multiple iterations, which impose significant time and memory overhead for the training phase. To improve training efficiency, we introduce the following approximations:\\
\textcircled{\scriptsize{1}} For step ({\romannumeral2}), we just add random noise to it directly via diffusion forward process by Eq.~\eqref{addnoise}.
\begin{equation}
\label{addnoise}
z_{L} =\mathcal{A}(z_0,L)=\sqrt{\alpha_L} z_0 + \sqrt{1 - \alpha_L} \epsilon, \quad \epsilon \sim \mathcal{N}(\mathbf{0}, \mathbf{I}),
\end{equation}
where $L$ is a hyperparameter in this paper with $0<L\leq T$ and $\mathcal{A}$ is the operator that transfers $z_0$ to its noisy version.\\
\textcircled{\scriptsize{2}} For steps ({\romannumeral3}) and ({\romannumeral4}) to obtain a clean edited image for calculating the classification loss (Eq.~\eqref{cls_loss}), we propose computing the loss on the decoded image from $\hat{z}_L$ rather than $\hat{z}_0$. To prevent the classifier from handling noisy data \(p(a\mid\mathcal{D}(\hat{z}_{L}))\), we make the following approximation:
\begin{align}\label{approx}
p(a\mid \mathcal{D}(\hat{z}_{L})) &=\mathbb{E}_{\hat{z}_0 \sim p(\hat{z}_0 \mid \hat{z}_{L})} \left[ p(a \mid \mathcal{D}(\hat{z}_{0})) \right] \nonumber\\
&\simeq p\left(a \mid \mathcal{D}\left(\mathbb{E}_{\hat{z}_0 \sim p(\hat{z}_0\mid \hat{z}_{L})}\left[\hat{z}_0\right]\right)\right), 
\end{align}
where $\mathbb{E}_{\hat{z}_0 \sim p(\hat{z}_0\mid \hat{z}_{L})}\left[\hat{z}_0\right]=\frac{\hat{z}_L - \sqrt{1 - \alpha_L} \Tilde{\epsilon}_{\theta}(\hat{z}_L,e_a)}{\sqrt{\alpha_L}}$.\\
Now the decoded image $\hat{x}$ in Eq.~\eqref{cls_loss} can be:
\begin{align}\label{predx0}
    \hat{x}=\mathcal{G}(x,e_a)=\mathcal{D}\left(\frac{\hat{z}_L - \sqrt{1 - \alpha_L} \Tilde{\epsilon}_{\theta}(\hat{z}_L,e_a)}{\sqrt{\alpha_L}}\right).
\end{align}
Notably, Eq.~\eqref{predx0} still allows us to incorporate the conditional guidance $e_a$.\\

\begin{figure}[tb]
    \centering
    \includegraphics[width=0.99\columnwidth]{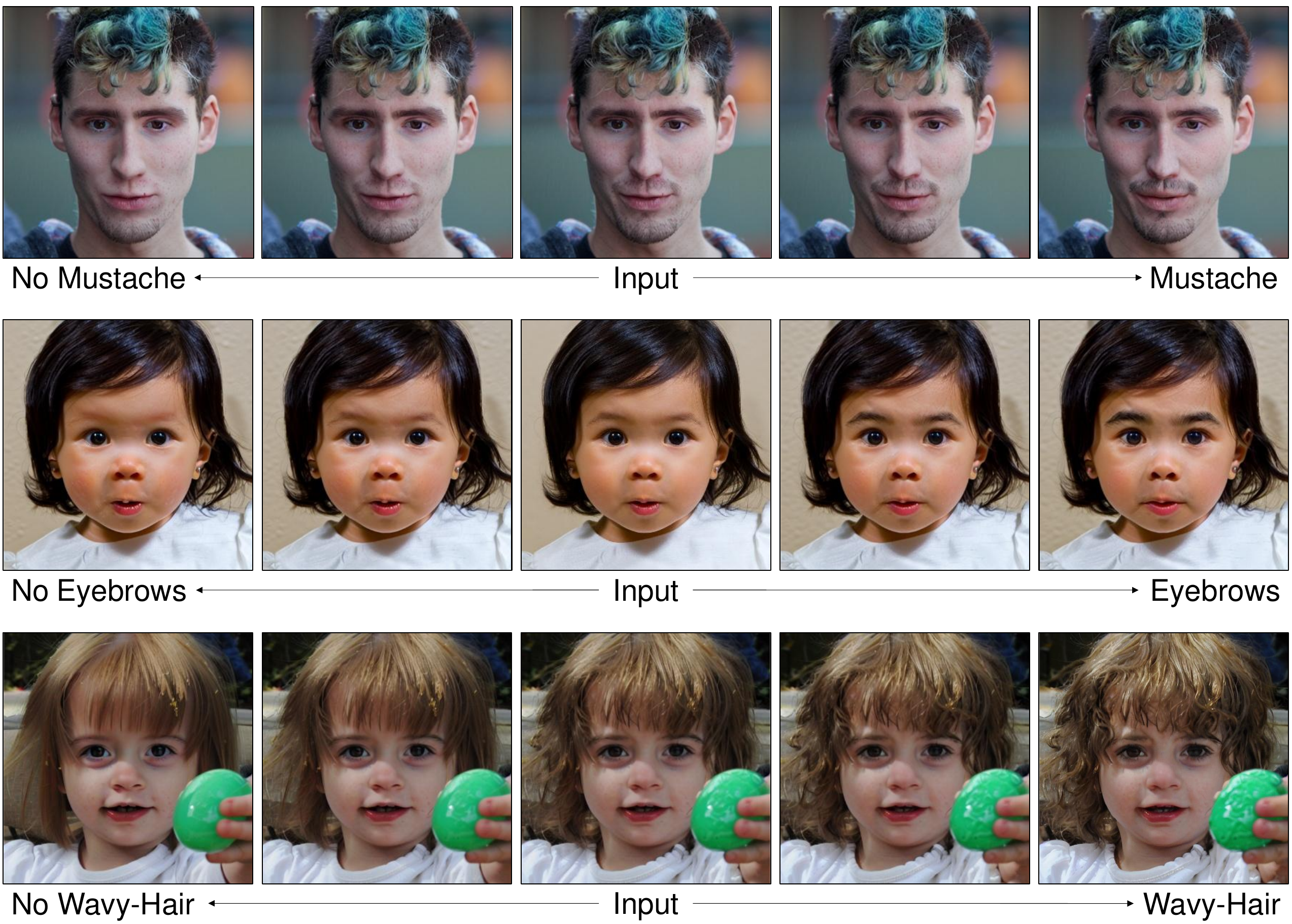} 
    \caption{\textbf{CASO Interpolation Results.} Our method allow
     users to implement fine-grained editing and bidirectional editing by simply  changing the classifier free guidance scale.}
    \label{interpolation}
\end{figure}

\begin{remark} 
The approximation error of Eq.~\eqref{approx} can be quantified with the Jensen gap \cite{DPS}. Its upper bound is:
\begin{equation}
\label{gap}
\mathcal{J} \leq \frac{d}{\sqrt{2\pi\sigma^2}} e^{-1/2\sigma^2} \left\| \nabla_{z} \mathcal{F}(\mathcal{D}(z))\right\| \mathbf{Q},
\end{equation}
where $d$ is environmental dimension (dimension of $z$), $\sigma^2$ is the ambient noise when the classifier predicts the label. And when $\sigma\rightarrow0$, $\frac{d}{\sqrt{2\pi\sigma^2}} e^{-1/2\sigma^2}\rightarrow0$. $\|\nabla_{z} \mathcal{F}(\mathcal{D}(z))\| := \max_{z^{\prime}} \|\nabla_{z^{\prime}} \mathcal{F}(\mathcal{D}(z^{\prime}))\|$  and $\mathbf{Q} := \int \|z_0 - \hat{z}_0\| p(z_0|z_L) \, dz_0.$  $\|\nabla_{z} \mathcal{F}(\mathcal{D}(z))\| $ is generally finite \cite{DPS} and when we choose the value of $L$ carefully, the upper bound on $\mathcal{J}$ can be sufficiently small. When $L$ is too large, $\|z_0 - \hat{z}_0\|$ in $\mathbf{Q}$ becomes excessively large, reducing classifier accuracy. Conversely, when $L$ is too small,  $z_L$ is close to clean data, making it difficult to edit the image at this time. 
\end{remark}

\paragraph{Reconstruction loss}
To preserve the attribute excluding details, we add a latent reconstruction loss:
\begin{equation}
    \mathcal{L}_{\text{rec}} = \mathbb{E}_{z,a}\left[\ell_{\text{r}}\left(\hat{z}_0, z_0\right)\right].
\end{equation}
where $\ell_{\text{r}}$ is the reconstruction loss. The final training objective is as follows:
\begin{equation}
\label{final_loss}
    \mathop{\min}_{\{e_a\}_{a=1}^K}\mathcal{L}_{\text{edit}}+\gamma \mathcal{L}_{\text{rec}}.
\end{equation}
We show the impact of this part of the loss function on the results in Appendix.
Our training algorithm is Algorithm.\ref{algorithm}.

\begin{table}
    \centering
    \begin{tabular}{lc@{\hspace{6pt}}c@{\hspace{6pt}}c@{\hspace{6pt}}c}
        \toprule
        Method & cat$\to$dog & dog$\to$cat & cat$\to$tiger & tiger$\to$cat \\
        \midrule
    Comp-Diff & 0.48 & 0.59 & 0.59 & 0.44 \\
    SEGA & 0.57 & 0.50 & 0.56 & 0.55 \\
    Cycle-Diff & \textbf{0.40} & \underline{0.48} & \underline{0.47} & \underline{0.40} \\
    Ours & \underline{0.43} & \textbf{0.42} & \textbf{0.45} & \textbf{0.34} \\
        \bottomrule
    \end{tabular}
    \caption{Comparison of various methods with LPIPS($\downarrow$) for animal type transfer.}
    \label{tab:comparison_afhq}
\end{table}

\section{Experiment}
To demonstrate the advancement and generalization of our method, we perform the following experiments.
\subsection{Experimental Setup} We use Stable Diffusion-v1.5\footnote{https://huggingface.co/stable-diffusion-v1-5/stable-diffusion-v1-5} for all experiments following \cite{dalva2024noiseclr}. The datasets used include: FFHQ \cite{FFHQ}, AFHQ \cite{AFHQ}, CelebAHQ \cite{CelebaHQ} and Stanford Cars datasets \cite{cars}. For more details, please refer to Appendix. 

Timesteps are also critical for editing quality and disentanglement \cite{uncovering,hertz2022prompt}. During training process, $L$ is set to $0.3T$ for human face and $0.4T$ for others. During editing, for subtle features like eyebrows, we start to apply our direction from $t\in[0.1T,  0.3T]$, while for some coarse-grained changes like species, editing at earlier timesteps is required ($t\in[0.8T,  0.9T]$). The results we show in the main text are all done with timesteps $T=50$. Our classifiers are all obtained by simply fine-tuning a VGG16 model \cite{vgg}. With a well-trained classifier, only 100-200 images are needed to train the embedding.

We compare our method with: DiffAE \cite{diffusionautoencoders}, Cycle-Diff \cite{cyclediffusion}, SEGA \cite{SEGA}, Comp-Diff \cite{compositional}, NoiseCLR \cite{dalva2024noiseclr} and T-LOCO\footnote{LOCO is purely unsupervised editing method. We found it is hard to obtain specified attribute editing directions for comparison. So we adopt its text extension version (T-LOCO) for the experiments.} \cite{loco}.

\subsection{Qualitative and Quantitative Results}
\begin{figure}[tb]
    \centering
    \begin{subfigure}[b]{0.45\textwidth} 
        \centering
        \includegraphics[width=\textwidth]{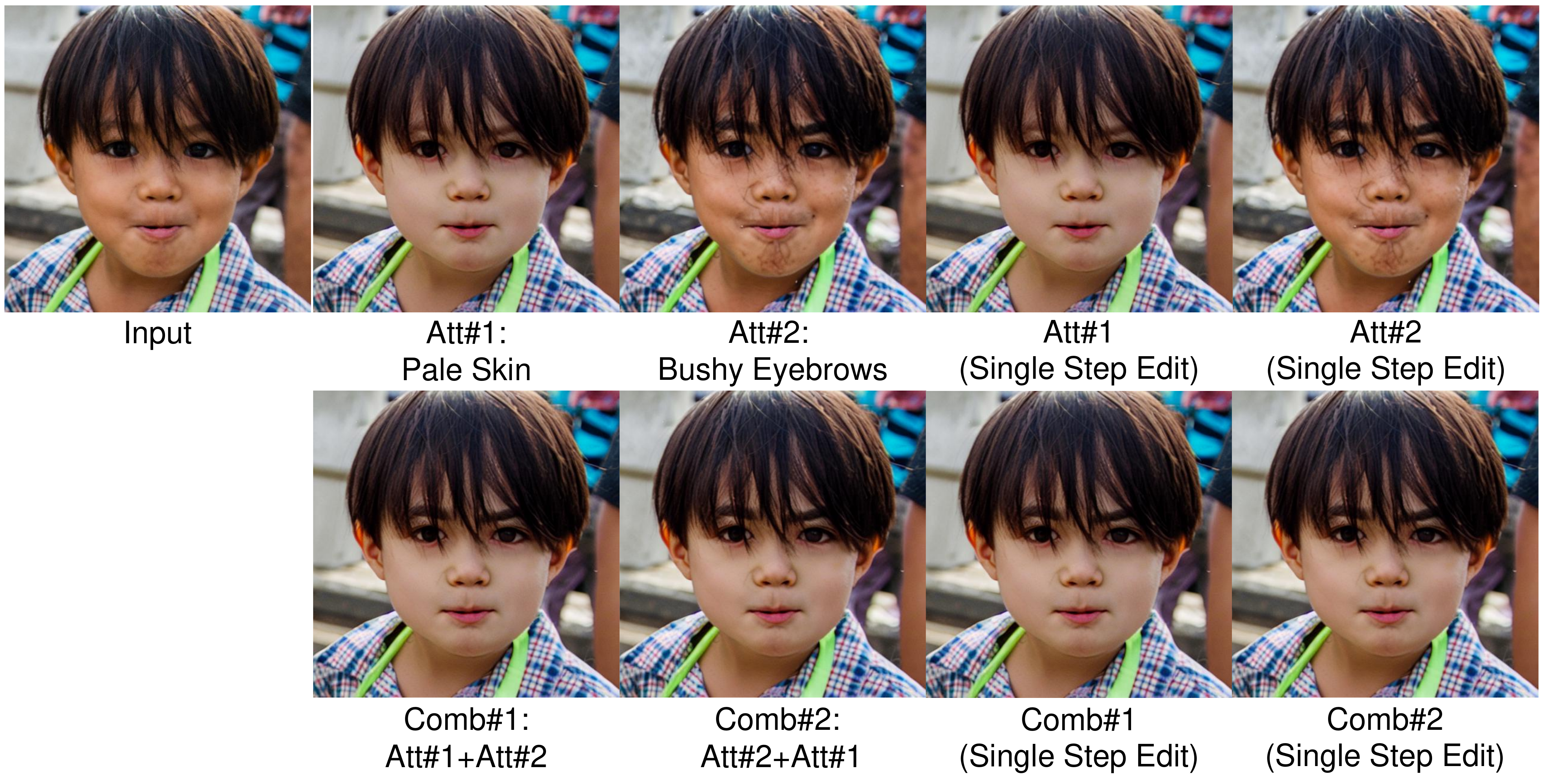} 
        \caption{Muti-attribute edit result. We compare the effect of the position order of the embedding stitching (the second row) and whether single-step editing is used (the last two images in each row are single-step edited) ) on the editing effect. When target area ``eyebrow" is occluded by hair, and our method can still complete the editing well.} 
        \label{muti a}
    \end{subfigure}
    \hfill 
    \begin{subfigure}[b]{0.45\textwidth}
        \centering
        \includegraphics[width=\textwidth]{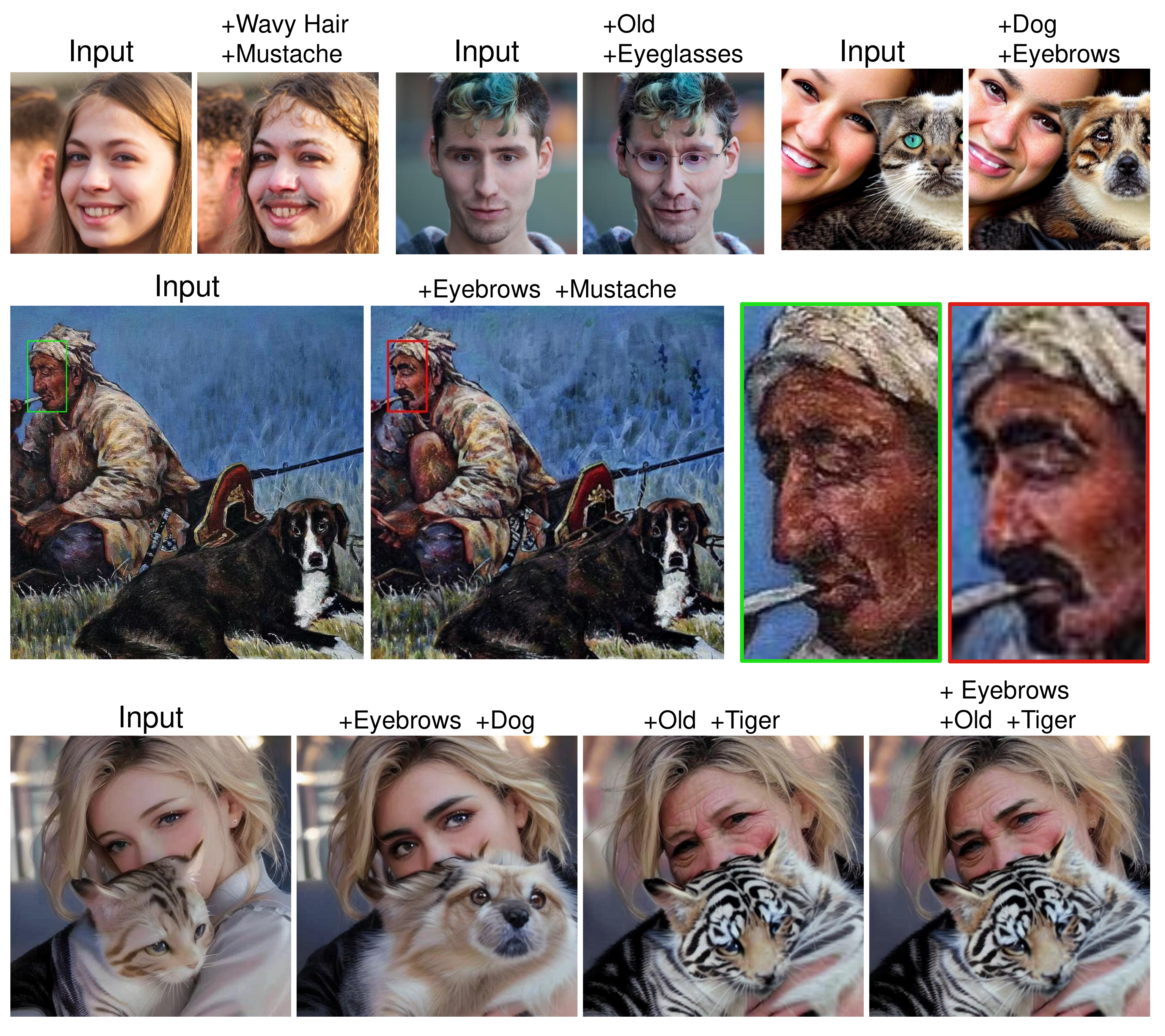}  
        \caption{Multi-attribute edits in a cross-domain context. The last input image for "woman and cat" in the first column is generated by SD.}
        \label{muti b}
    \end{subfigure}
     \caption{\textbf{CASO Muti-attribute Edit}. In complex and challenging scenarios, our method can still achieve perfect editing.}
     \label{muti}
\end{figure}

Our qualitative results are shown in Fig.~\ref{edit}. Although our approach is trained on real world images, it exhibits remarkable generalization on images of other styles, demonstrates that our method generalizes beyond previous works.  Notably, our edits are not confined to the data seen by the classifier. For example, if we use a well-trained dog and cat classifier learn semantic embedding for ``cat", it enables edits not only for cat and dog, but also all of others. The results demonstrate that our embeddings can capture precise semantics for the target attribute.

We compare the generalization performance with other methods for attribute ``Mustache".  This is shown in Fig.~\ref{compare}. Regardless of the type of image, our model consistently achieves superior performance.

Table~\ref{tab:comparison} and  Table~\ref{tab:comparison_afhq} shows quantitative results.
We calculate the LPIPS metric (lower is better) \cite{lpips} for editing different attributes on human face data and species conversion between animal data\footnote{For pecies conversion between animal data, we do not compare our method with NoiseCLR, DiffAE and T-LOCO. This is because  NoiseCLR and DiffAE do not support such editing. And T-LOCO performs poorly on this structure changes greatly task and the statistical results are not meaningful. We show examples in the Appendix.}. Our method  achieve lower LPIPS than the other methods, indicating greater coherence while performing the edits.

\begin{figure}[tb]
    \centering
    \begin{subfigure}[b]{0.45\textwidth}  
        \centering
        \includegraphics[width=\textwidth]{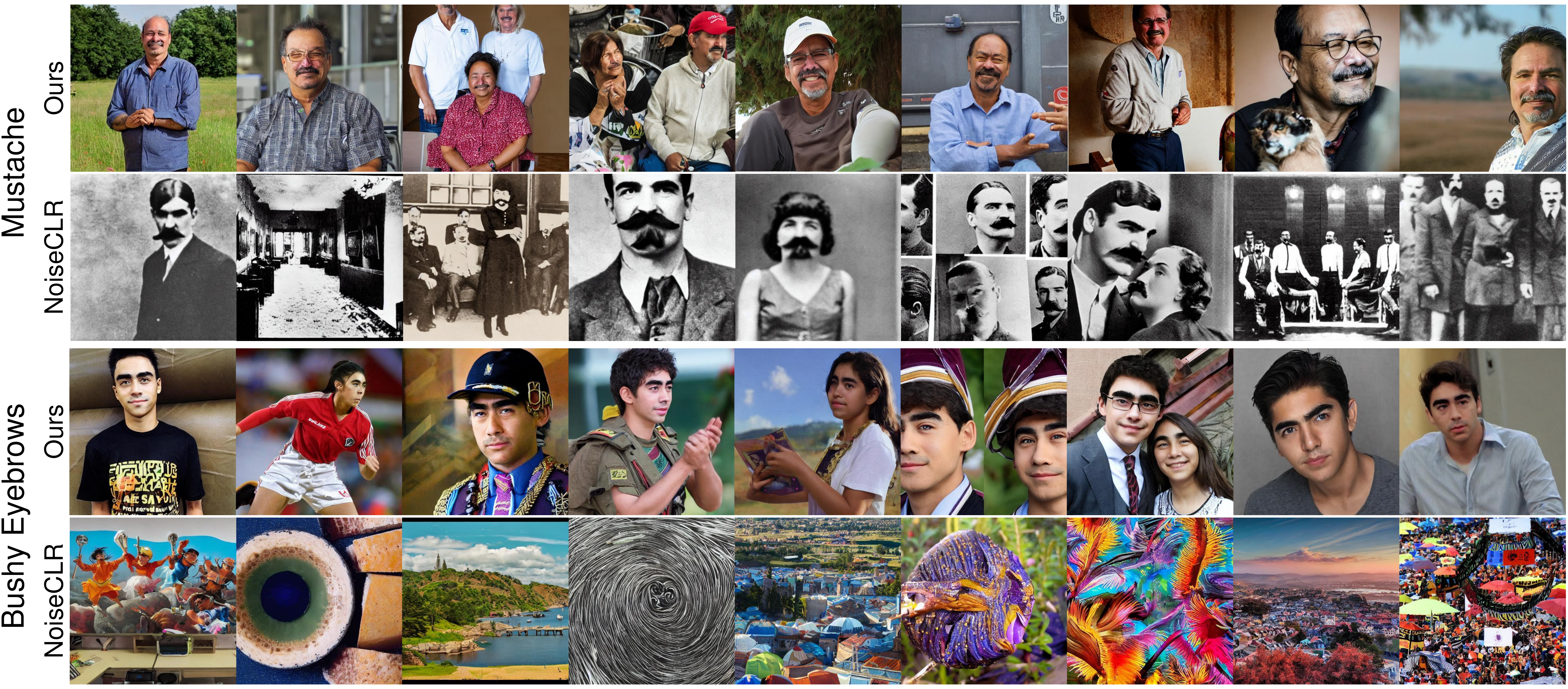} 
        \caption{Sample from Gaussian noise with different condition guidance.}
        \label{fig:sub1}
    \end{subfigure}
    \hfill 
    \begin{subfigure}[b]{0.45\textwidth}
        \centering
        \includegraphics[width=\textwidth]{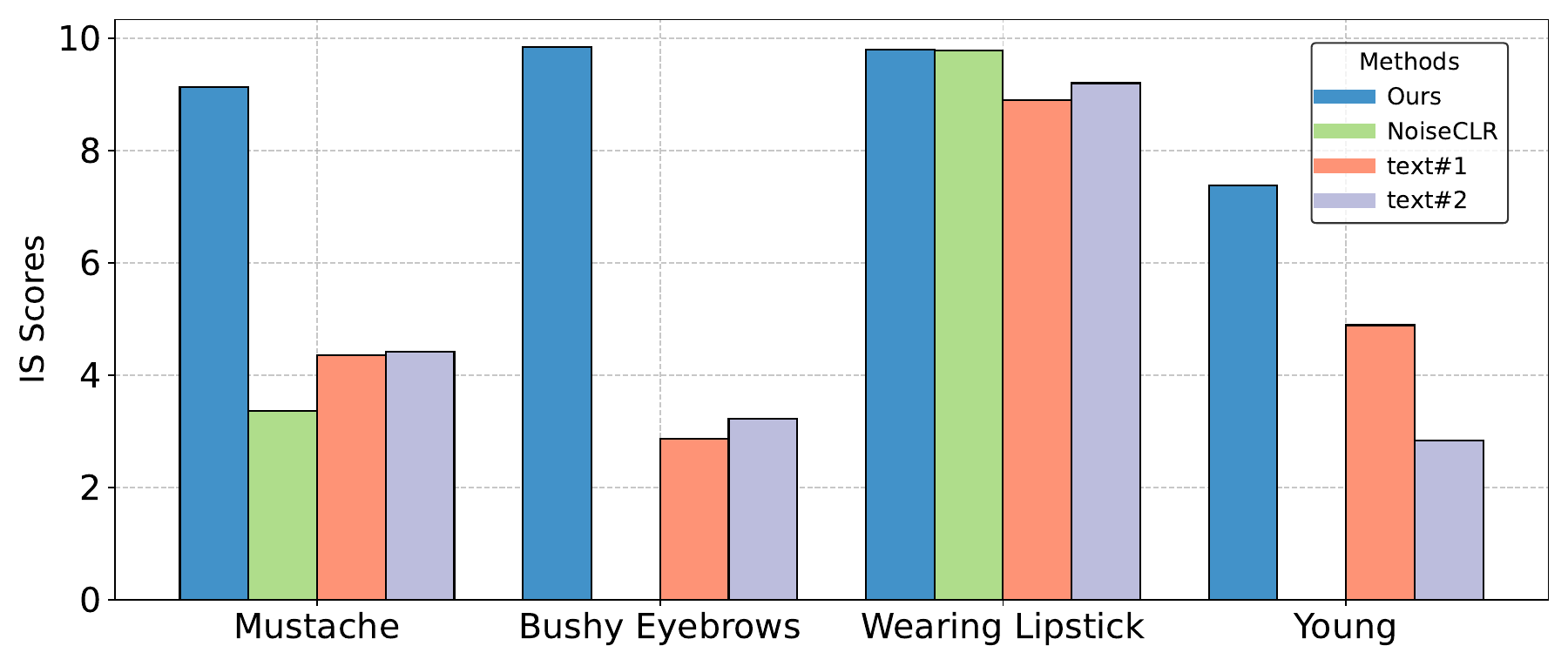}  
        \caption{Image quality of generation under different guidance, measured by IS ($\uparrow$). We do not count the IS score for some directions of NoiseCLR since it is not a desirable editing direction or the authors do not open source weights.}
        \label{fig:sub2}
    \end{subfigure}
    
    \caption{\textbf{Generation Result.} We use different direction: learned from NoiseCLR, learned from ours, and two kinds of text prompt (see Appendix) to guide the generation of images from Gaussian noise with corresponding attributes and calculate their IS scores.}
    \label{sample}
\end{figure}
\noindent \textbf{Interpolating Edit} Our approach can achieve fine-grained attribute editing and reverse editing by simply changing the classifier free guidance scale $\lambda$. For instance, using the ``Bushy Eyebrows" embedding, users can create the effect of ``Sparse Eyebrows" by setting the $\lambda$ to a negative value.  Some examples are shown in Fig. \ref{interpolation}. A similar effect can be achieved by changing the time at which the condition embedding is added, but this does not allow for reverse editing. We show it in the Appendix.

\noindent \textbf{Multi-attribute Edit} Our method allows for multi-attribute editing  by simply concatenating multiple embeddings. Embedding available on different datasets can be combined (e.g. human face and animal). Experiments in Fig. \ref{muti a} show that the position order of embedding stitching does not affect the result of image editing, which also proves the semantic decoupling of the learned orientation. We show editing results in difficult scenarios: In Fig. \ref{muti a}, the target region to be edited has occlusion. In Fig. \ref{muti b} middle, an artwork image, the face region to be edited is only a small part of the whole image. These are complex scenarios where our model still performs the editing task perfectly. 

\noindent \textbf{Single-Step Edit} To further verify the effectiveness and consistency of our method, we conduct a single-step editing experiment, as shown in Fig.~\ref{muti a}. In this experiment, guidance information is applied in only one step of the denoising process, while all other steps before and after rely on unconditional embeddings. In contrast, most existing methods require conditional guidance across multiple denoising steps.
Experimental results in Fig.~\ref{muti a} show that our outcomes are highly consistent, regardless of whether single-step editing is used. This demonstrates that our editing method is powerful enough that images generated by a single-step editing already exhibit the target properties. 
\begin{table}
    \centering
    \begin{tabular}{lcccc}
        \toprule
        \multirow{2}{*}{\makecell{DDIM \\Steps}} & \multicolumn{2}{c}{AFHQ Cat} & \multicolumn{2}{c}{AFHQ Dog} \\
        \cmidrule(lr){2-3} \cmidrule(lr){4-5}
         & $\lambda=0$ & $\lambda=3$ & $\lambda=0$ & $\lambda=3$ \\
        \midrule
        $T=10$ & 85.39 & \textbf{39.00}\textcolor{red}{\scriptsize{$\downarrow54.3\%$}} & 88.52 & \textbf{52.03}\textcolor{red}{\scriptsize{$\downarrow41.2\%$}} \\
        $T=20$ & 70.34 & \textbf{32.41}\textcolor{red}{\scriptsize{$\downarrow53.9\%$}} & 47.19 & \textbf{45.13}\textcolor{red}{\scriptsize{$\downarrow4.37\%$}}  \\
        $T=50$ & 72.35 & \textbf{43.20}\textcolor{red}{\scriptsize{$\downarrow40.4\%$}} & \textbf{50.55} & {50.89}\textcolor{blue}{\scriptsize{$\uparrow0.67\%$}} \\
        \bottomrule
    \end{tabular}
    \caption{Reconstruction quality on AFHQ w/o or w/ guidance, measured by FID ($\downarrow$). $\lambda=0$ means reconstruction w/o guidance.}
    \label{FID_result}
\end{table}

\subsection{Accurate Semantics}\label{generation}
\begin{figure}[htb]
    \centering
    \begin{subfigure}[b]{0.48\columnwidth} 
        \includegraphics[width=\textwidth]{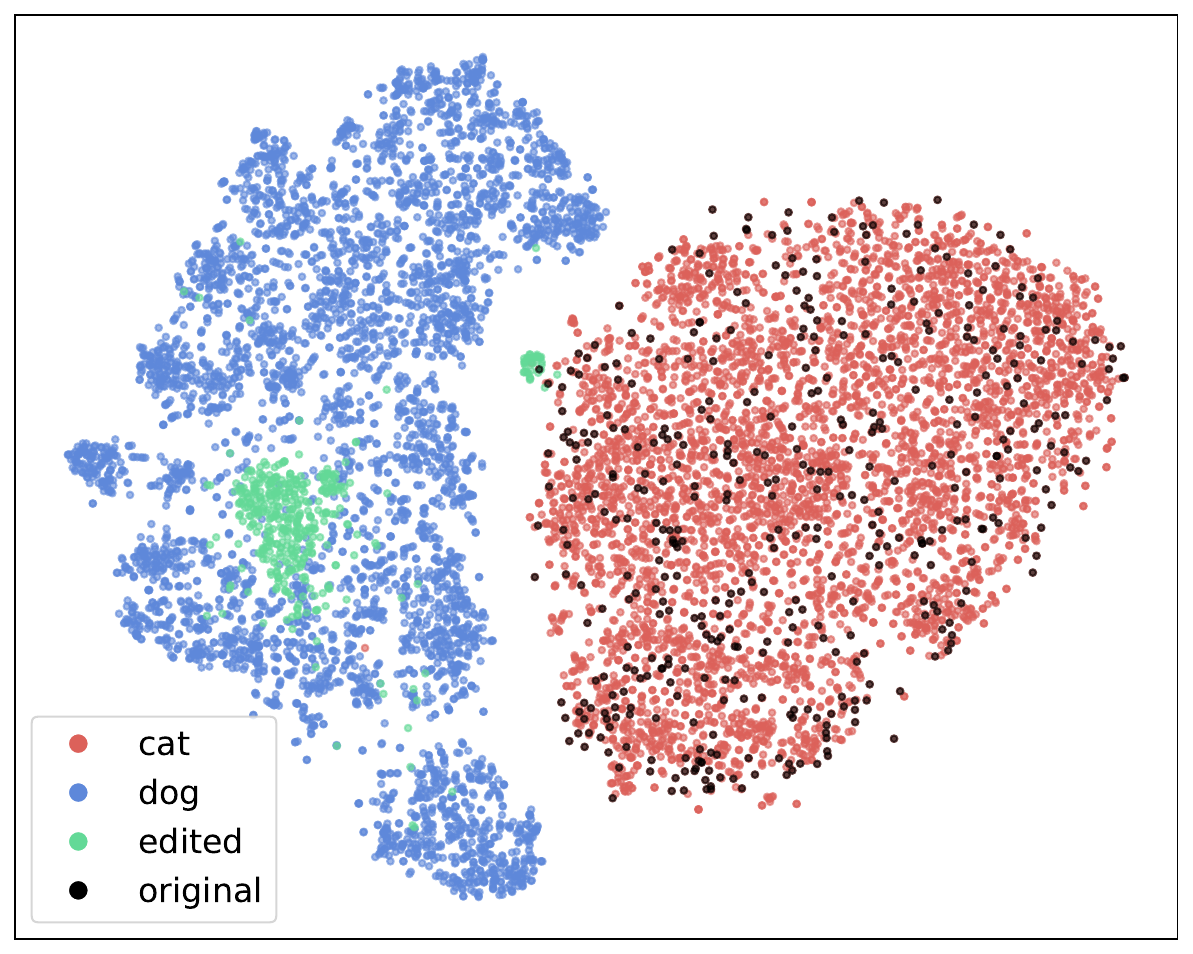}
        \caption{Ours}
    \end{subfigure}
    \hfill
    \begin{subfigure}[b]{0.48\columnwidth}
        \includegraphics[width=\textwidth]{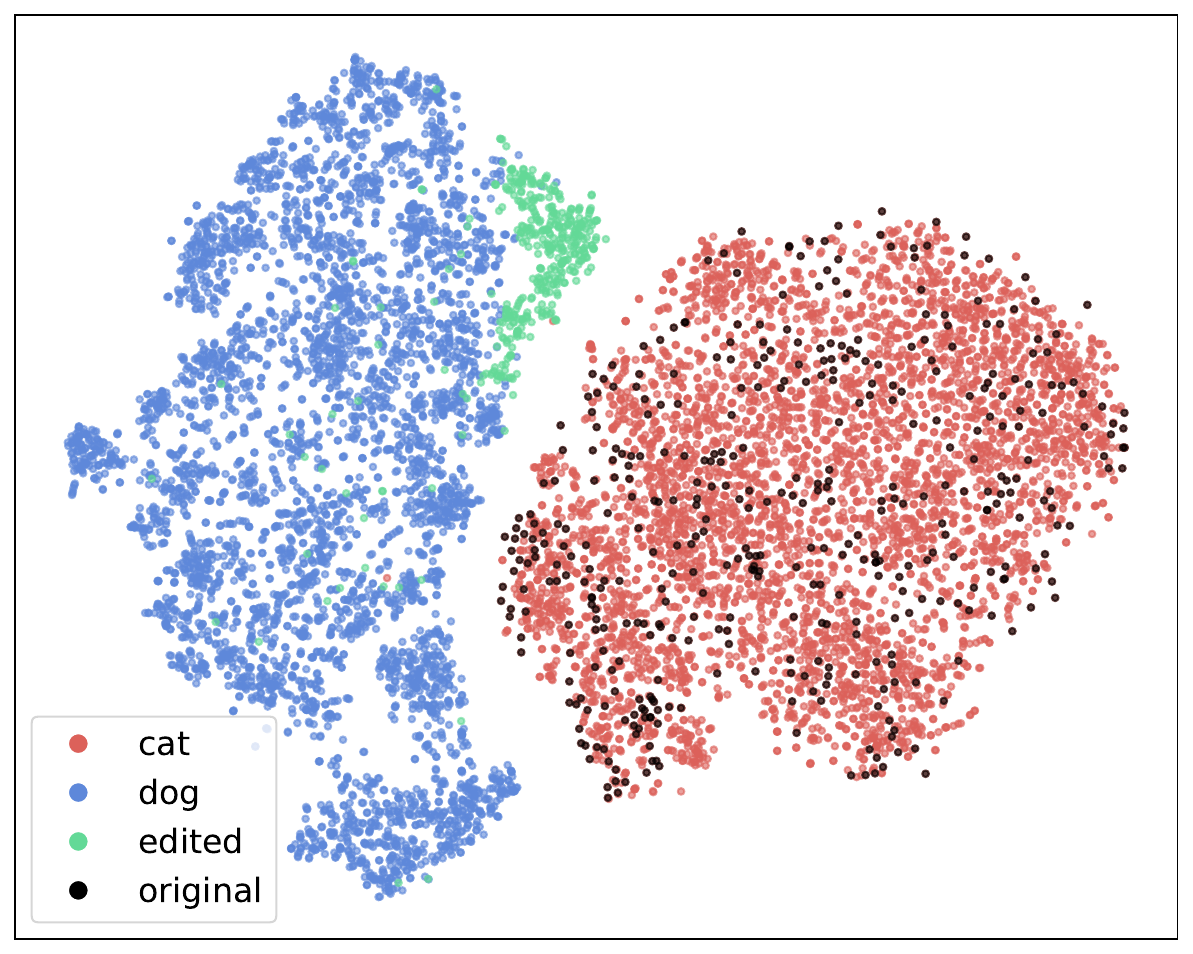}
        \caption{SEGA}
    \end{subfigure}
    \caption{T-SNE of cat$\rightarrow$dog image edit. Red dot: real cat images; blue dots: real dog images; black dots: real cat images randomly sampled for editing; green dots: the generated images (to be edited as dog from the sampled cat images).} \label{tsne}
\end{figure}

To demonstrate that our model can learn the most precise semantic direction, we directly sample from Gaussian noise with different guidance (ours, NoiseCLR and text). The results are shown in Fig.~\ref{fig:sub1}. The generated images demonstrate that our learned embeddings more accurately capture semantic features. For instance, comparing the generated images guided by ``Bushy Eyebrows" learned using our method and NoiseCLR separately, the faces generated by our method consistently exhibit the ``Bushy Eyebrows" property, while the images generated by NoiseCLR do not have the desired semantics. We attribute the ability of NoiseCLR to perform attribute editing despite learning incorrect semantics to the coincidence of feature subspaces in the data space, which is also mentioned in SEGA \cite{SEGA}. Additionally, we evaluate the IS (Inception Score) \cite{is} of different guided sampling images (higher is better), which is shown in Fig. \ref{fig:sub2}. Some IS scores are not reported, because it is not an ideal edit direction (Fig.~\ref{fig:sub1}) or the authors do not open source its weight~\footnote{We tried to reproduce it. But because the direction of convergence obtained by such unsupervised methods is not controllable, we did not get the direction of interest based on their method.}. Experiments show that our guidance can effectively improve the quality of image generation.

We conduct another experiment in which we randomly sample from the cat dataset, edit the images into dogs using different methods, including our approach and SEGA \cite{SEGA} (the leading text-based image editing work at present). We visualize the features obtained by the classifier (i.e., $\mathcal{F}_{-1}(x)$) using T-SNE, as shown in  Fig.~\ref{tsne}. Notably, the features of images edited by our CASO locate at the dog class mean, consistent with our theoretical result. This demonstrate that our embedding indeed captures accurate attribute semantics at the dataset level, enabling superior editing performance.

\subsection{Not Only Edit} \label{Not Only Edit}
\begin{figure}[tb]
    \centering
    \includegraphics[width=0.45\textwidth]{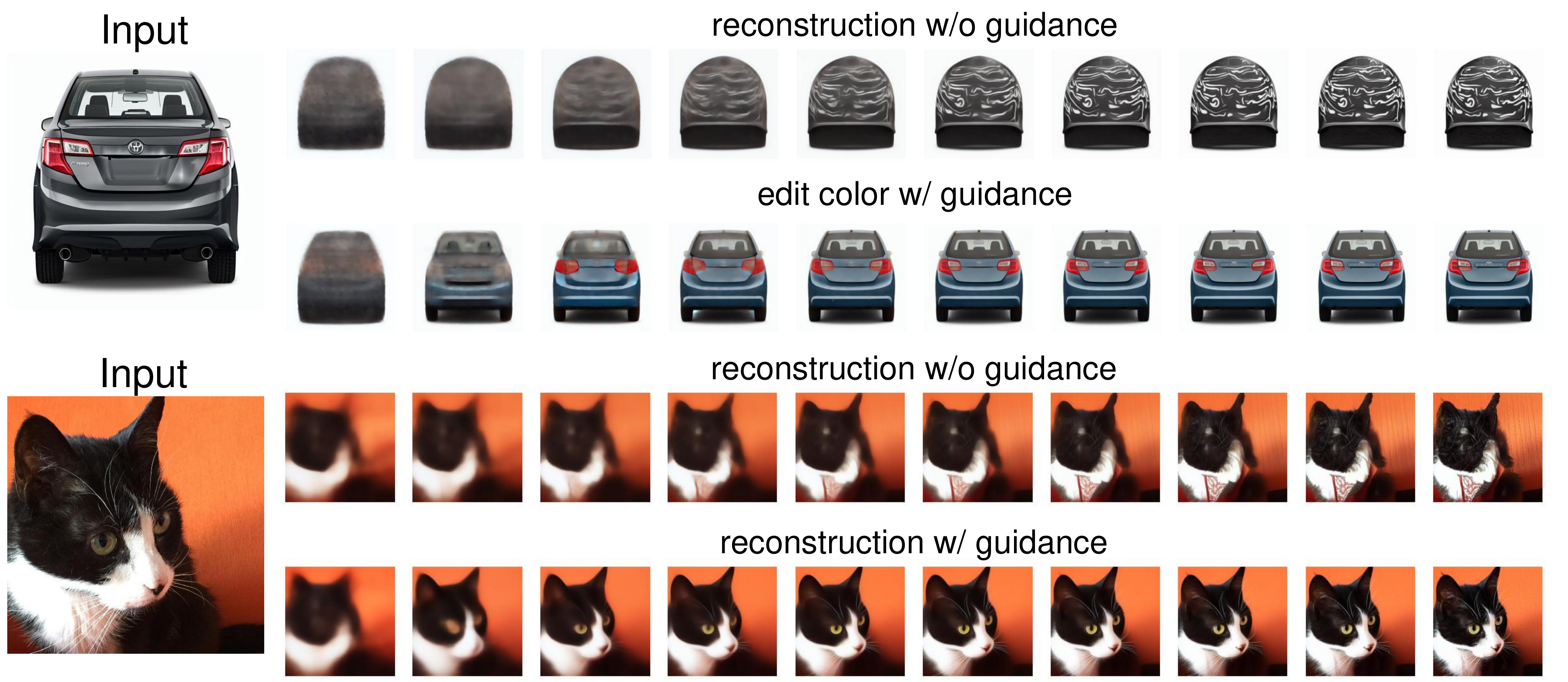} 
    \caption{\textbf{Improve image quality during editing and reconstruction.} Our guidance can not only faithfully reconstruct the structure of the original image, but also achieves the editing of the target attribute.}
    \label{improve}
\end{figure}

Many editing works based on pre-trained diffusion models struggle to achieve satisfactory results due to the model's limited ability to accurately reconstruct certain images. In our experiments, we demonstrate that our method not only enables effective image editing but also enhances reconstruction quality, as shown in Fig.~\ref{improve}. We argue that the embeddings learned by our approach effectively capture both the overall structural information of the edited object and its target attributes.
 
 When editing an image using attributes identical to those in the original (e.g., guiding the denoising of a cat image with the embedding of “cat”), the task effectively becomes reconstruction. This leads to higher-quality results and faster convergence. As shown in Fig.~\ref{improve}, our method enhances image quality in tasks such as car color editing and cat image reconstruction. This also explains the high quality of the generated images in \ref{generation}. However, we acknowledge that an excessively large $\lambda$ may introduce distortions in the reconstructed image. More examples can be found in the Appendix.

We calculated the FID metrics \cite{fid} for AFHQ Cat and Dog datasets under unconditional reconstruction and guided reconstruction with ``cat" and ``dog" embeddings, respectively. The results are presented in Table \ref{FID_result} and Fig. \ref{improve}. Experiments show that our embeddings significantly enhance the quality of reconstruction for images within the same class.

\section{Conclusion} 
We propose an image editing framework that optimizes semantic embeddings with a few image inputs, guided by a classifier, to deliver high-quality, disentangled edits and superior generalization compared to existing methods. However, since it builds upon the pre-trained Stable Diffusion model, its effectiveness is limited by Stable Diffusion. Additionally, the potential for malicious misuse, such as deepfake creation, is a concern \cite{deepfakes}. We suggest implementing strict access controls, maintaining usage logs, and promoting responsible use to ensure ethical deployment.




\section*{Acknowledgments}
The work was supported by the National Key R\&D Program of China (2023YFC3306100), National Natural Science Foundation of China (62172324, 62272379, 62403429), National Research Foundation, Key R\&D in Shaanxi Province (2023-YBGY-269, 2022-QCY-LL33HZ), Xixian New Area Science and Technology Plan Project (RGZN-2023-002, 2022 ZDJS-001), Zhejiang Provincial Natural Science Foundation (LQN25F030008). 
This paper was also supported by the National Research Foundation, Singapore under its National Large Language Models Funding Initiative (AISG Award No: AISG-NMLP-2024-004). Any opinions, findings and conclusions or recommendations expressed in this material are those of the author(s) and do not reflect the views of National Research Foundation, Singapore.

\bibliographystyle{named}
\bibliography{ijcai25}

\clearpage

\appendix
\section{Proof of Proposition 2} 
First, we introduce some additional theoretical results from~\cite{neuralcollapse,papyan2020prevalence} to lay the foundation for the proof of Proposition~2.
\begin{definition}\label{notation}
 The linear classifier can be represented by weights $W=[w_1, w_2, \ldots, w_K]^T \in\mathbb{R}^{K\times p}$, where $K$ is the number of classes, $p$ is the dimension of the linear classifier. We define the between-class covariance as:
 \begin{equation}
     \Sigma_\mathrm{B} = \mathbb{E}_a \left[\mu_a\mu_a^\top \right],
 \end{equation}
 the within-class covariance:
 \begin{equation}
     \Sigma_\mathrm{W}=\mathbb{E}_{i,a}\left[(h_{i,a}-\mathbb{E}_{i}[h_{i,a}])(h_{i,a}-\mathbb{E}_{i}[h_{i,a}])^\top\right],
 \end{equation}
 the train total covariance:
 \begin{equation}
     \Sigma_\mathrm{T}=\mathbb{E}_{i,a}\left[(h_{i,a}-\mathbb{E}_{i,a}[h_{i,a}])(h_{i,a}-\mathbb{E}_{i,a}[h_{i,a}])^\top\right],
 \end{equation}
where $h_{i,a}$ and $\mu_{a}$ is defined in Definition~1 of the main text.
Recall from multivariate statistics that \cite{papyan2020prevalence}:
 \begin{equation}
     \Sigma_\mathrm{T}=\Sigma_\mathrm{B}+\Sigma_\mathrm{W}
 \end{equation}
\end{definition}

\begin{theorem}[\cite{neuralcollapse,papyan2020prevalence}]\label{nc2} When the classifier is well-trained, $\Sigma_\mathrm{W}\rightarrow0$ and $\Sigma_\mathrm{T}=\Sigma_\mathrm{B}$. The optimal classifier weights are given by: 
\begin{equation}\label{weight}
    W=\frac{1}{K}M^\top\Sigma_\mathrm{T}^\dagger=\frac{1}{K}M^\top\Sigma_\mathrm{B}^\dagger,
\end{equation}
where $\dagger$ denote the Moore-Penrose pseudoinverse, and $M=[\mu_1,\dots,\mu_K]$. The features $\mu_{a}$ convergence to a simple geometric structure called an Equiangular Tight Frame (ETF) :
\begin{subequations}
\begin{align}
& \frac{\langle \mu_a, \mu_{a'} \rangle}{\|\mu_a\|_2 \|\mu_{a'}\|_2} \rightarrow 
    \begin{cases}
1, & a = a' \\
\frac{-1}{K-1}, & a \neq a'
\end{cases} \\
& \|\mu_a\|_2 - \|\mu_{a'}\|_2 \rightarrow 0 \quad \forall a \neq a'
\end{align}
\end{subequations}
\end{theorem}
\noindent \underline{The proof for Proposition~2} is as follows:
\begin{proof}
First, we define
\begin{equation} \label{sigma-b}
     \Sigma_\mathrm{B}'(e_a) = \mathbb{E}_a \left[\mu_a'(e_a)(\mu_a'(e_a))^\top \right].
 \end{equation}
Let $M(e_a)=[\mu_1'(e_1),\dots,\mu_K'(e_K)]$ denote the stack matrix of class mean.  According to Theorem~\ref{nc2}, when Eq.~\eqref{cls_loss}(refer to the main text), we can obtain
\begin{equation}\label{b}
    W = \frac{1}{K} M(e_a)^\top (\Sigma_\mathrm{B}'(e_a))^\dagger.
\end{equation}
Eq.~\eqref{sigma-b} implies:
\begin{equation}\label{a}
    \Sigma_\mathrm{B}'(e_a)=\frac{1}{K}M(e_a) M(e_a)^\top
\end{equation}
By combining Eq.~\eqref{b} and Eq.~\eqref{a}, we get:
\begin{equation}
    W=M(e_a)^\top(M(e_a) M(e_a)^\top)=M(e_a)^\dagger.
\end{equation}
Theorem~\ref{nc2} implies that $M(e_a)$ has exactly $K-1$  no-zero and equal singular values, so $M(e_a)=\beta M(e_a)^\dagger$ for some constant $\beta$ \cite{papyan2020prevalence}. Thus, we can obtain:
\begin{equation}
    W=\beta M(e_a),
\end{equation}
Therefore, for each classifier weight $w_a$, there exists a correspondence between the class mean $\mu_a'(e_a)$ and the classifier weight: $w_a=\beta\mu_a'(e_a)$. It can be reformulated in the following form:
\begin{equation}
    \frac{w_{a}}{\left\|w_{a}\right\|_{2}}-\frac{\mu^{\prime}_{a}(e_a)}{\left\|\mu^{\prime}_{a}(e_a)\right\|_{2}}\rightarrow 0.
\end{equation}
The proof is complete.
\end{proof}


\section{Direct text guidance is disentangled}
\begin{figure}[htb]
    \centering
    \includegraphics[width=0.95\columnwidth]{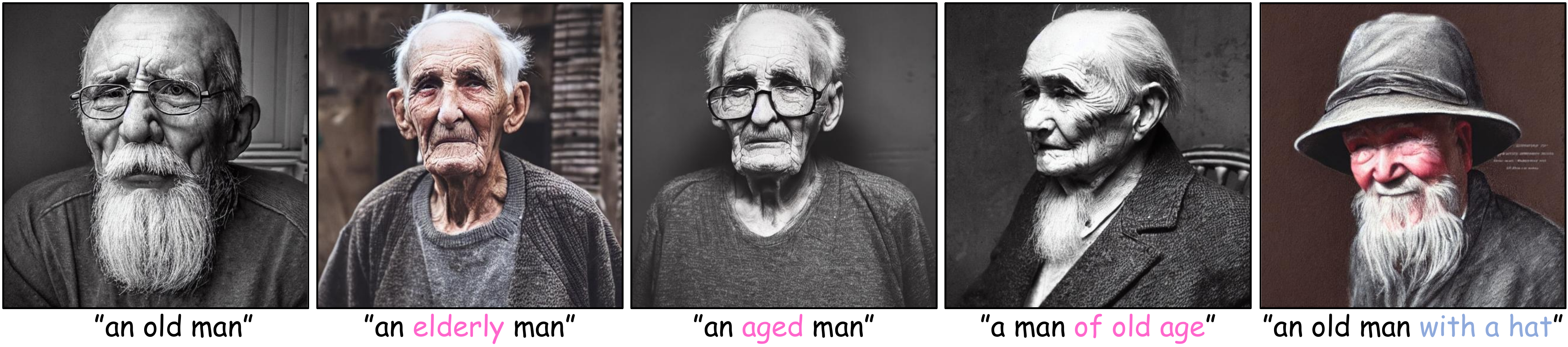} 
    \caption{Sample results generated with the same random seed but different text prompts.}
    \label{text generation}
\end{figure}

\section{Problem with blurred images after editing}
In some of the images in the main text, it can be seen that the edited images may be somewhat blurry (such as Fig.~\ref{fig:sub2}). This is due to the poor reconstruction effect of SD on these pictures. We are merely demonstrating the universality of our method in complex situations. The reconstructed image is shown below.
\begin{figure}[htb]
    \centering
    \includegraphics[width=0.95\columnwidth]{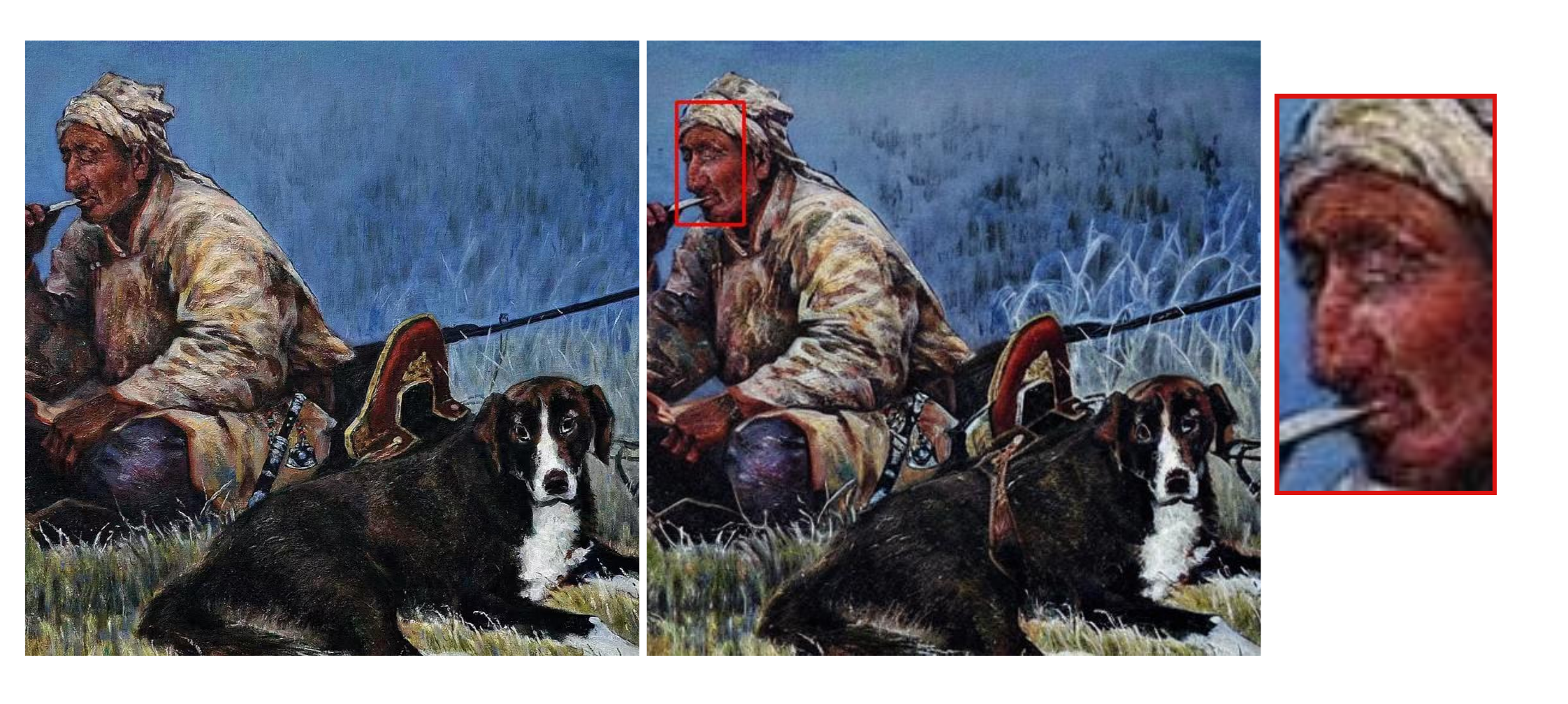} 
    \caption{The reconstruction results of Fig.~\ref{fig:sub2} (On the far left is the original image).}
\end{figure}

\section{Experiment Details} \label{Experiment details}
\subsection{Dataset}
We use the CelebAHQ \cite{CelebaHQ}, FFHQ \cite{FFHQ}, AFHQ \cite{AFHQ}, and Stanford Cars datasets \cite{cars}. We train classifiers on CelebaHQ, AFHQ and Stanford Cars datasets by fine-tuning the VGG16 model. To be specific, we use 50\% of the data to train the face attribute classifiers on CelebaHQ, and train embeddings on FFHQ with 100-200 images.  For animal data, we train classifiers with training set of AFHQ (the dataset itself has been partitioned in advance) and approximately 200 images from each category were randomly selected for training semantic embeddings.
 For Stanford Cars dataset, we classify cars based on their type (labeled data) and train classifiers and embeddings.
 
 We use AdamW optimizer \cite{adamw} with learning rate $10^{-2}$ without warmup to train all semantic embeddings.  For the classification loss and reconstruction loss in Eq.~\eqref{final_loss} we both use the MSE loss.  And the default classifier free guidance scale $\lambda$ is set to 10.
 \subsection{Accurate Semantics}
 Here we show more details of the experiment in \ref{generation}.
\subsubsection{Text We Used in \ref{generation}}
In \ref{generation}, we compare pictures generated by different guidance, which include our method, NoiseCLR, and two kinds of text guides. The text we use is shown in Table~\ref{tab:text}.
\begin{table}[ht]
    \centering
    \begin{tabular}{lcc}
        \toprule
       Attribute & text\#1 & text\#2 \\
        \midrule
        Mustache & Mustache &\thead {People with \\ Mustache} \\
        Bushy Eyebrows & Bushy Eyebrows & \thead {People with \\ Bushy Eyebrows} \\ 
        Wearing Lipstick & Wearing Lipstick &\thead { People \\Wearing Lipstick} \\
        Young  & Young & Young People  \\
        \bottomrule
    \end{tabular}
    \caption{Text prompt we use in \ref{generation}.}
    \label{tab:text}
\end{table}
\subsubsection{IS Statistics In \ref{generation}}
In \ref{generation}, we generate images using different guidance information (ours, NoiseCLR and two kinds of text prompts) and counted their IS scores. However, we do not count the metrics of NoiseCLR on some attributes. This is because the semantic information they learned was wrong or the author did not open source the weights. 

The basic idea of NoiseCLR is to randomly initialize a number of directions (100 by default) and then use contrastive learning to continuously optimize these directions until they converge to a mutually decoupled state. These directions are labeled manually after convergence. However, this approach is extremely unstable, as the training result will depend on the variance of the dataset, etc. And the direction of convergence is not controllable, we can not control its convergence to a specific semantic. We reproduced the method but did not get the specific semantics, so this part of the index was not counted.

\subsection{Embedding Design}
\begin{figure}[htb]
    \centering
    \includegraphics[width=\columnwidth]{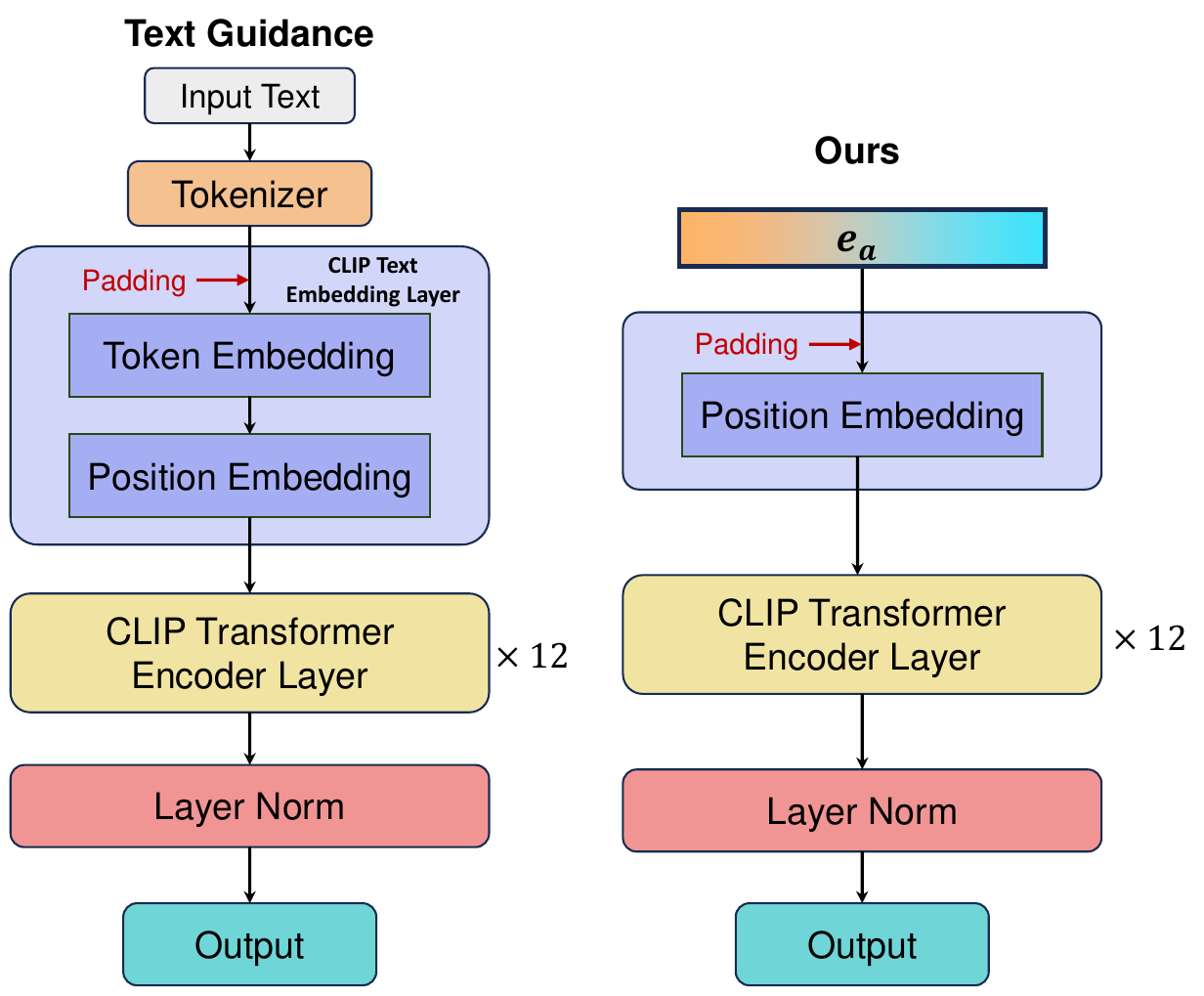} 
    \caption{Details about how our embedding are injected into the UNet to guidance edit.} \label{details}
\end{figure}
 Here we detail how traditional text prompt and our embedding are injected into the UNet network to guide the generation (Fig.~\ref{details}). For ``Padding" in the Fig.~\ref{details}, traditional text guidance uses end\_idx 49407 to fill to a fixed length and than then passes through Token Embedding Layer. In our approach, we directly use the embedding corresponding to end\_idx 49407 in the Token Embedding to populate the input.

\section{T-LOCO Edit}

\begin{figure*}
    \centering
    \includegraphics[width=0.8\textwidth]{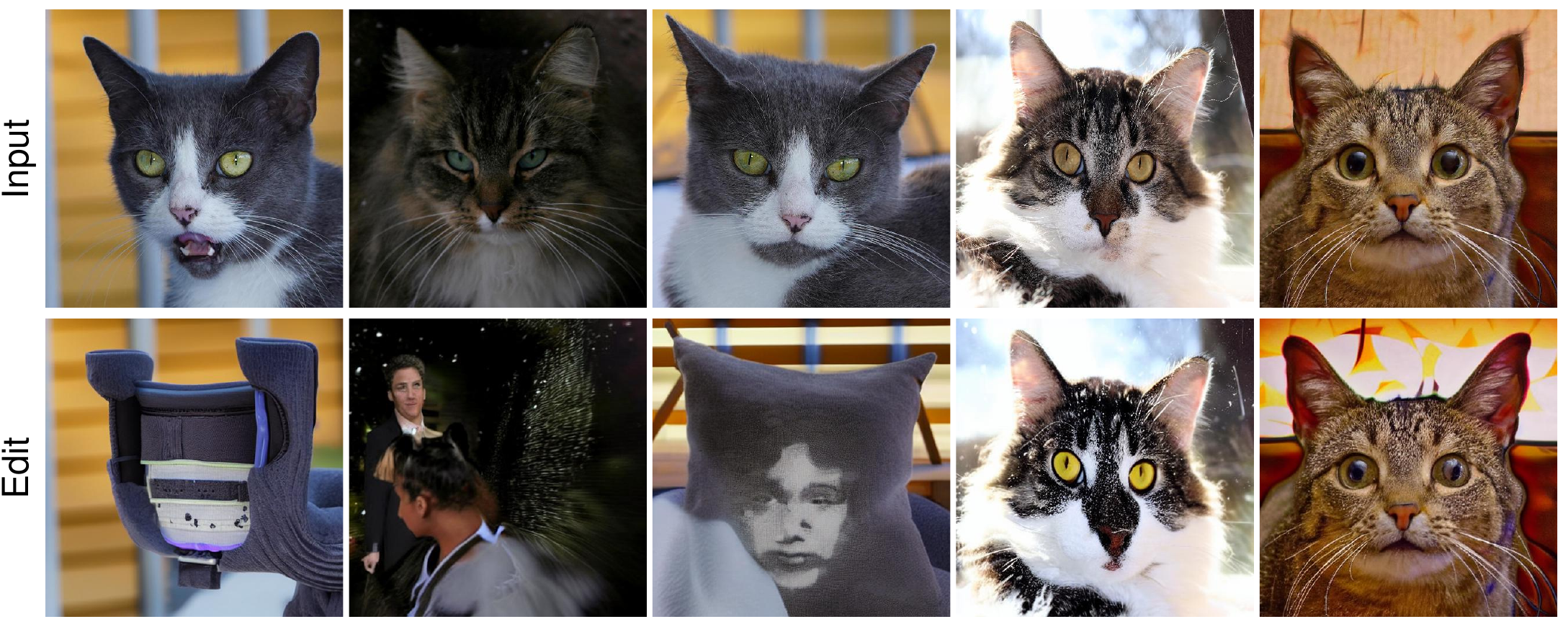} 
    \caption{T-LOCO edit for cat$\rightarrow$dog. Because of its poor effect, we do not conduct statistics of the relevant results.}
\label{t_loco}
\end{figure*}

We mentioned in the main text that T-LOCO performs poorly on edits with large structural changes, so we did not count the results of T-LOCO on animal species transfer edits.Here we show some examples in Fig.~\ref{t_loco}.

\section{Ablation Experiment}

\subsection{Reconstruction Loss}\label{ablation_wore}
The loss function of our method consists of two parts: the edit loss and the reconstruction loss.
When training without reconstruction loss, training is sometimes unstable because the model may converge to a distribution other than the one modeled by the classifier. We present some extreme examples in Fig. \ref{ablation}.

\subsection{Start Edit Time}
\begin{figure*}[htb]
    \centering
    \includegraphics[width=0.9\textwidth]{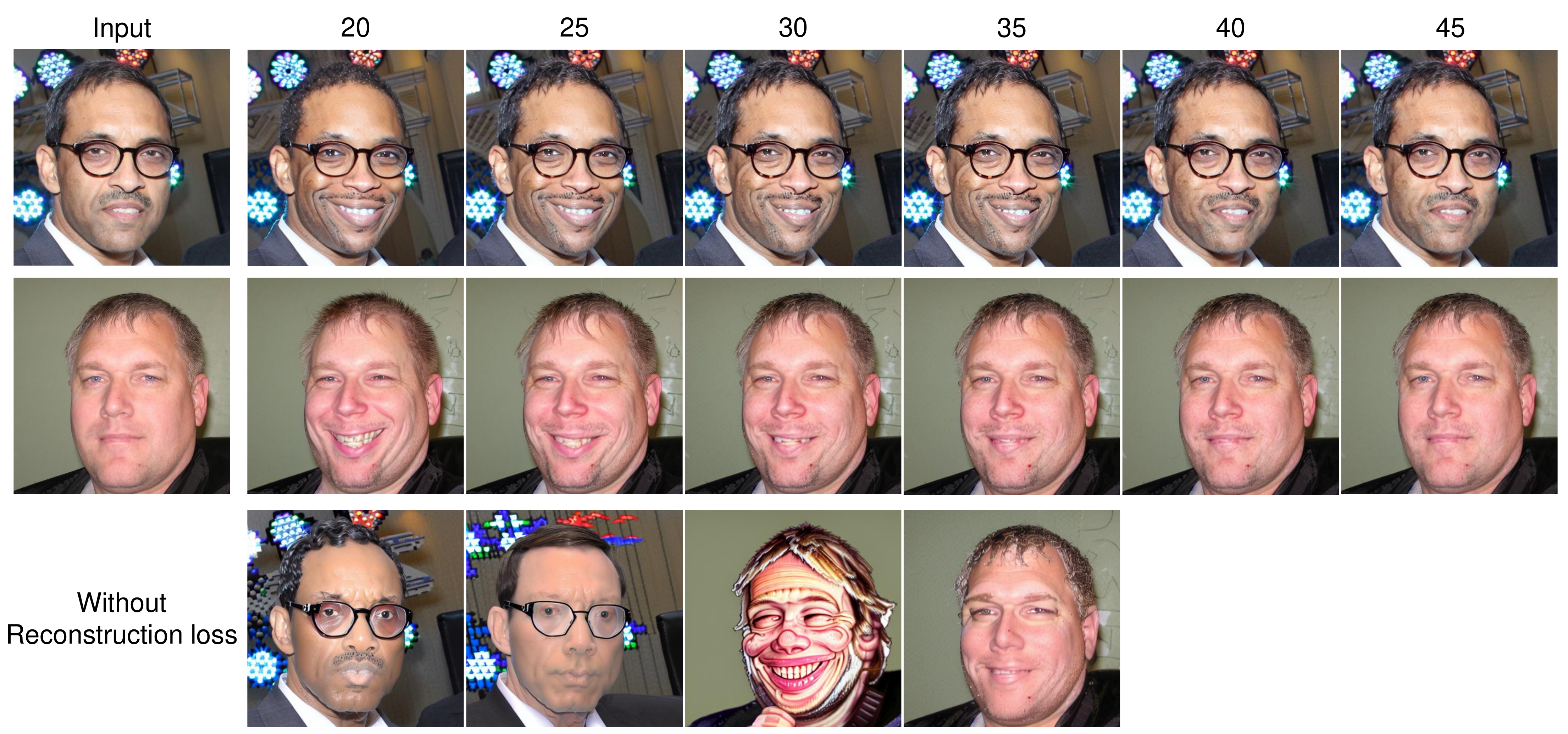} 
    \caption{With $T=50$, comparison of the effect of starting editing at different timesteps.}
    \label{ablation}
\end{figure*}

Because our method maintains a high degree of consistency between single-step edits and continuous edits, only the start edit time and classifier free guidance scale $\lambda$ affect the editing effect. We show the editing effect under different start guide edit times in Fig.~\ref{ablation} (for the ``Smile" attribute), with DDIM steps $T=50$. 

\subsection{Amount of Training Data}
\begin{figure}[htb]
    \centering
    \includegraphics[width=\columnwidth]{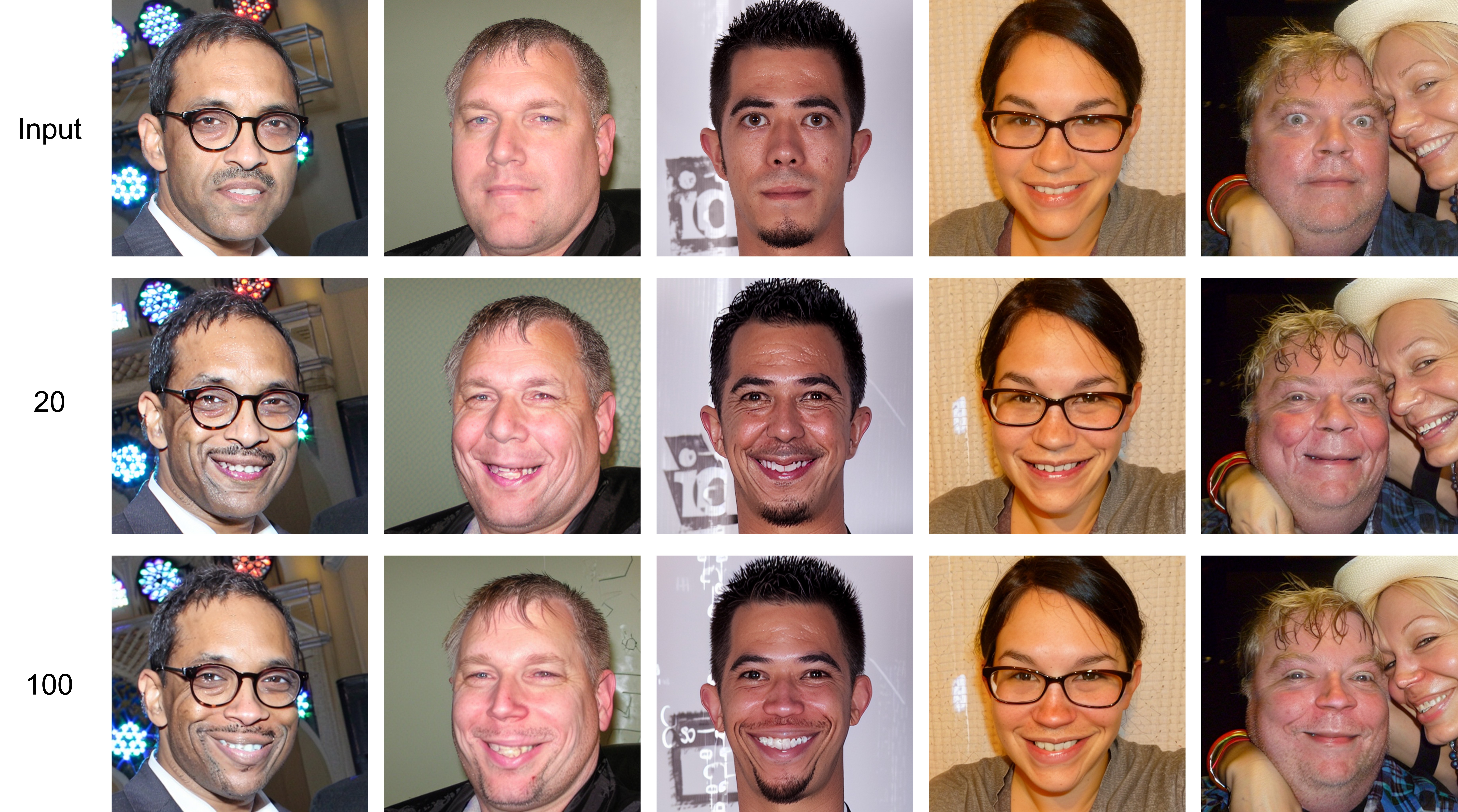} 
    \caption{Effect of using different numbers of images for training.}\label{ablation_num}
\end{figure}
We train our embeddings with different numbers of images, and we find that around 20 images are enough to edit, but when this number goes up to 100-200, the results are better, please see Fig.~\ref{ablation_num}.

\begin{figure*}[htb]
    \centering
    \begin{subfigure}[b]{\textwidth}  
        \centering
        \includegraphics[width=0.92\textwidth]{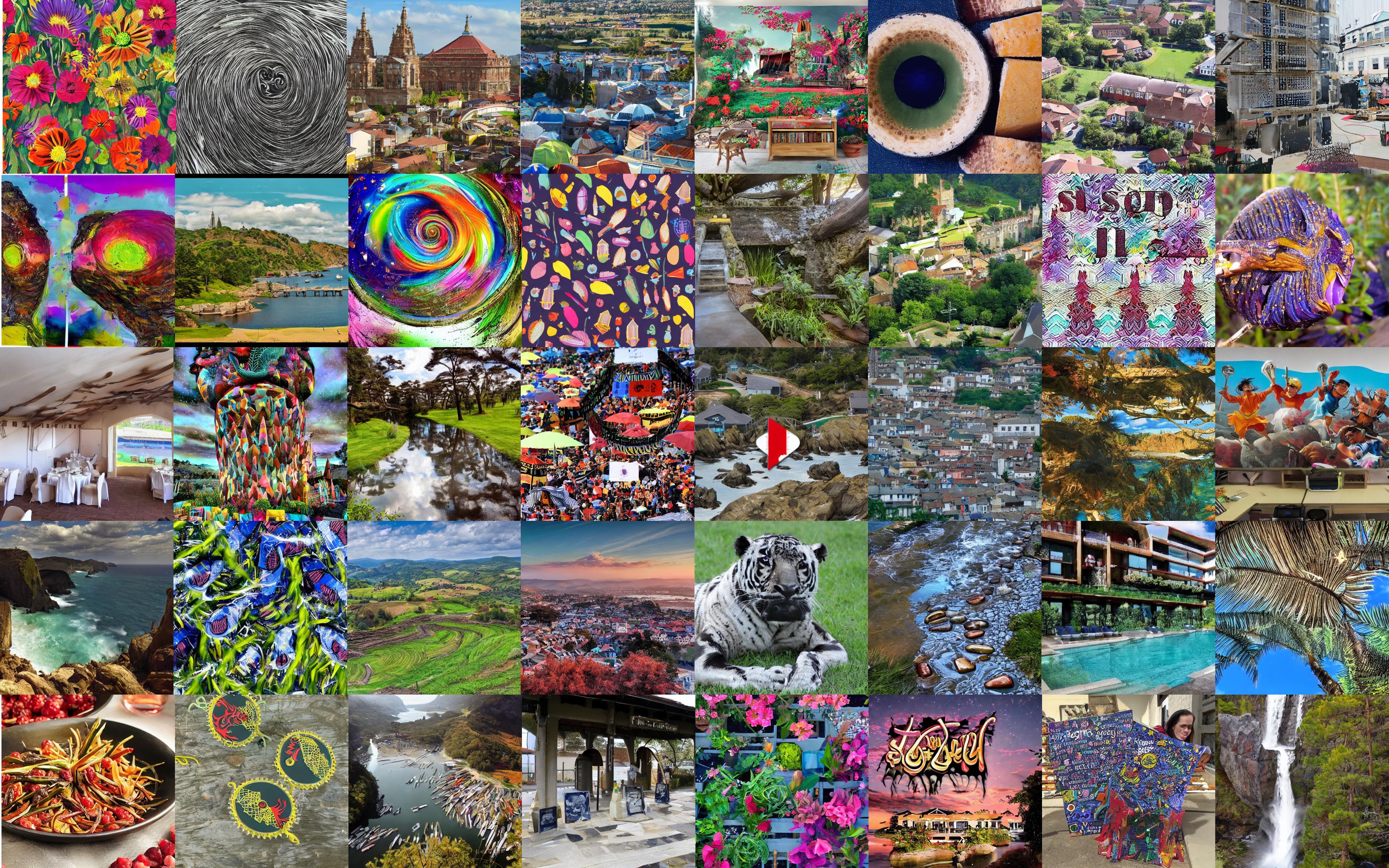}  
        \caption{NoiseCLR}
    \end{subfigure}
    \hfill 
    \begin{subfigure}[b]{\textwidth}
        \centering
        \includegraphics[width=0.92\textwidth]{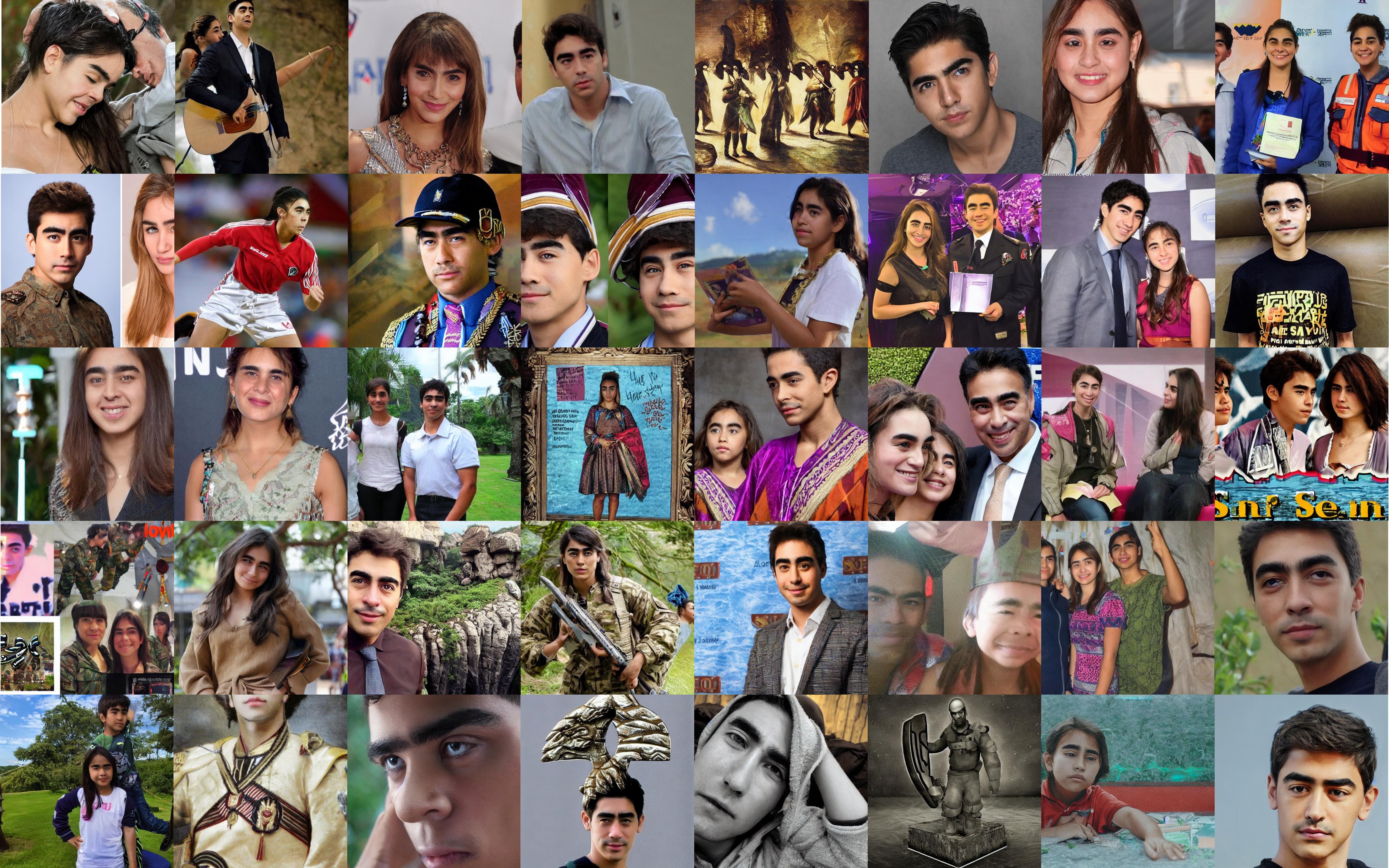}  
        \caption{Ours}
    \end{subfigure}
    \caption{Generation with different guidance for attribute ``Bushy Eyebrows".}\label{generation_eyebrow}
\end{figure*}

\section{Guidance Generation}\label{Guidance generation}
We show more results of sample from Gaussian noise with different guidance for attribute ``Bushy Eyebrows" (Fig. \ref{generation_eyebrow}). Our method generates faces with ``Bushy Eyebrows" features, whereas NoiseCLR doesn't learn the exact semantics.

\section{Not Only Edit}\label{ablation_not only edit}
\begin{figure*}[htb]
    \centering
    \includegraphics[width=\textwidth]{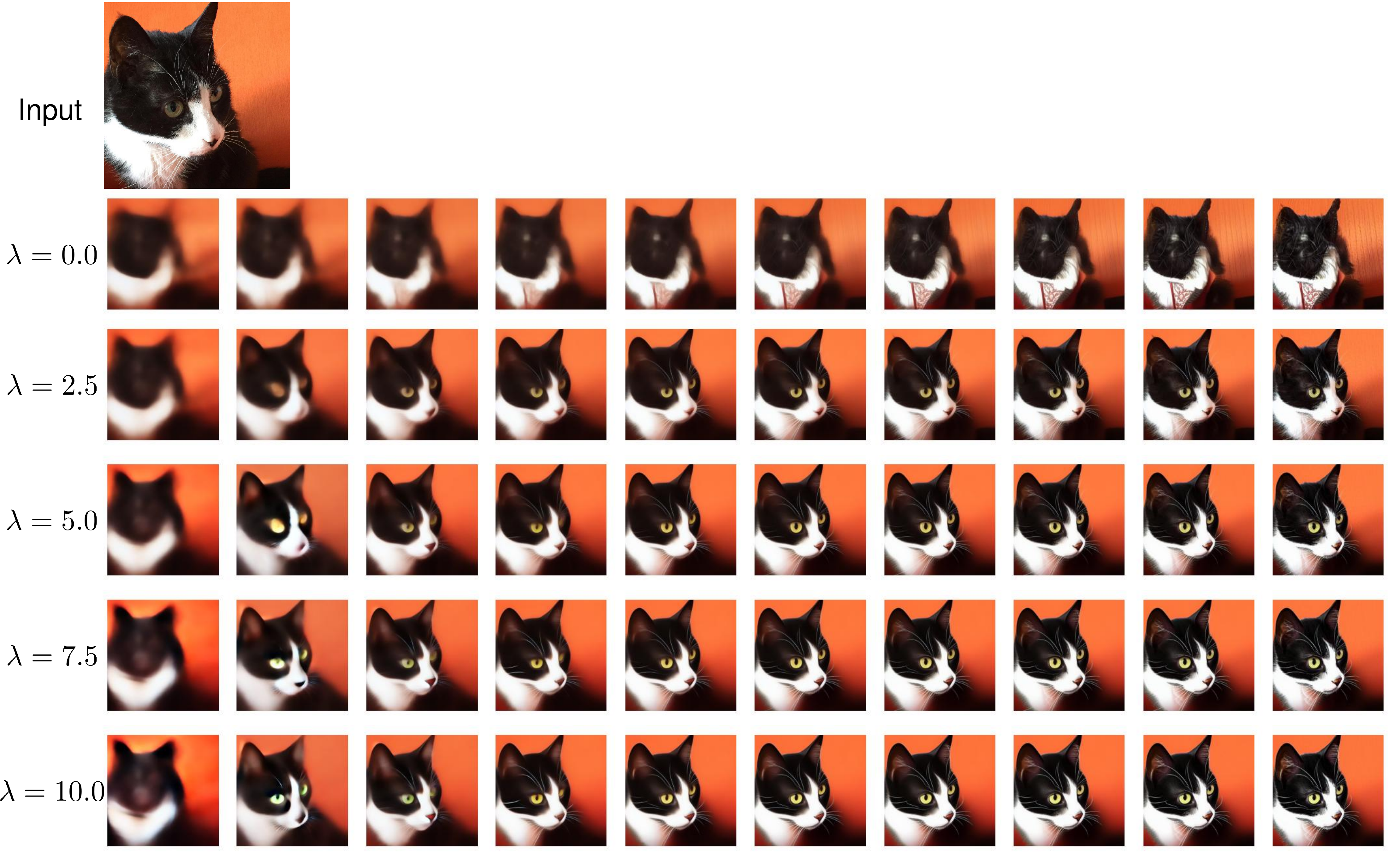} 
    \caption{Our embeddings can help in faster and better reconstruction of images of the same class.}
    \label{ablation_improve}
\end{figure*}

\begin{figure*}[htb]
    \centering
    \includegraphics[width=\textwidth]{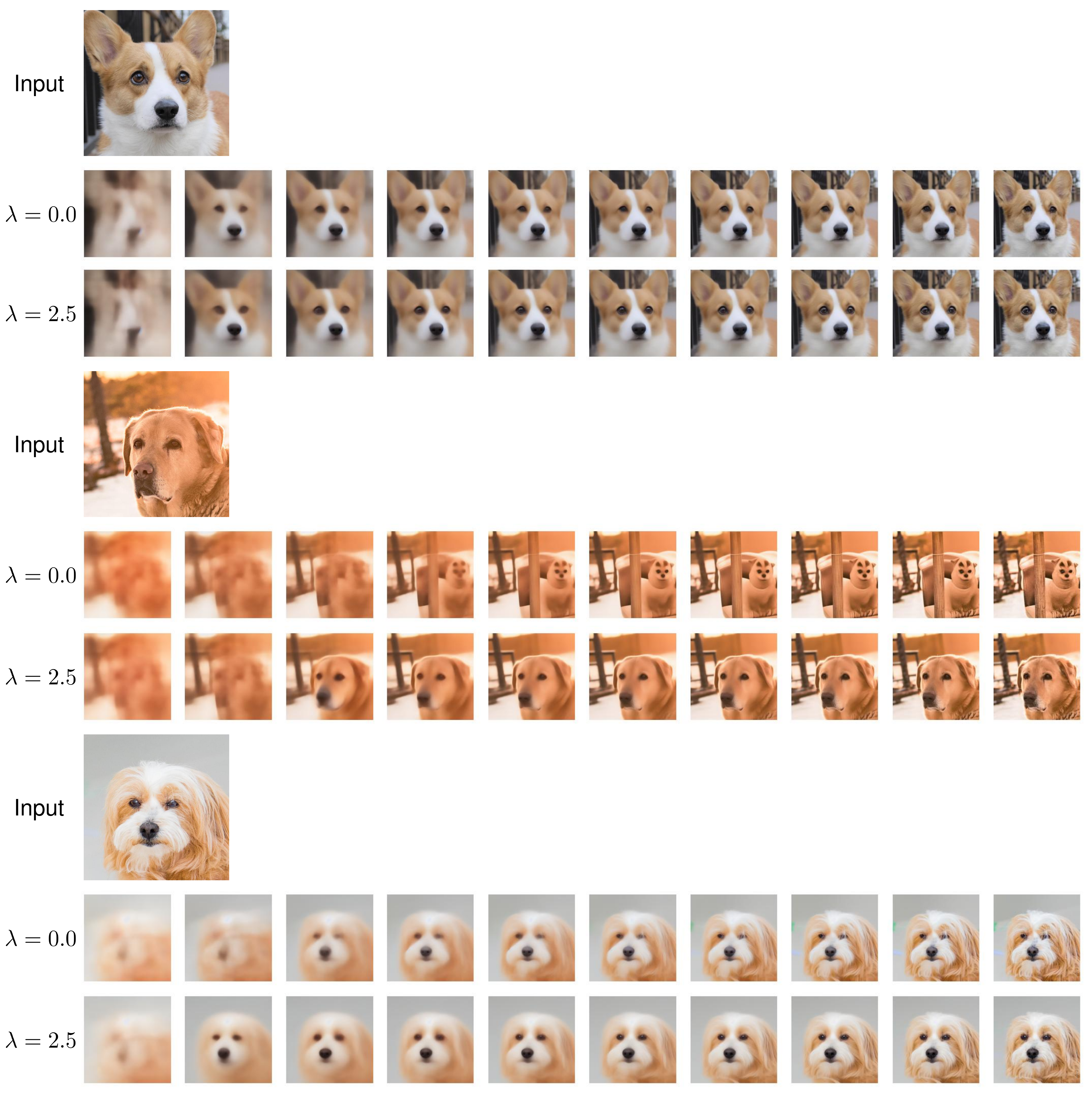} 
    \caption{Our embeddings can help in faster and better reconstruction of images of the same class.}
    \label{help_reconstruction_2}
\end{figure*}

We show the visual result of how our embeddings help with reconstruction at different classifier free guidance scales (Fig.~\ref{ablation_improve} and~\ref{help_reconstruction_2}). Note that smaller $\lambda$ is good when using our learned semantic embedding to guide the reconstruction. When $\lambda$ is too large, the image may be distorted.

\section{Images We used}
In addition to the datasets mentioned in the text, we also used some public images on the Internet in our research, thank them very much.
\begin{itemize}
\item \url{https://ixintu.com/sucai/0SkWkjqPk.html}
\item \url{https://www.amazon.sg/Ingres-after-Portrait-J4617-Poster/dp/B0BWCKKNYC}
\item \url{https://www.flickr.com/photos/91534998@N06/27779365535}
\item \url{https://www.arch2o.com/real-life-models-flora-borsi/}
\item \url{https://widgetopia.io/widget/anime-boys-widget-s}
\item \url{https://www.facebook.com/permalink.php/?story_fbid=470392145782887&id=100084363165671}
\item \url{https://today.line.me/tw/v2/article/PGyyV5R}
\item \url{https://openaccess.thecvf.com/content/CVPR2024/supplemental/Dalva_NoiseCLR_A_Contrastive_CVPR_2024_supplemental.pdf}
\item \url{https://www.mercedes-benz.com.sg/passengercars/models/cabriolet-roadster.html}
\item \url{https://www.facebook.com/login/?next=https%3A%2F%2Fwww.facebook.com%2F3012829978800166}
\item \url{https://changan.drive.place/hunter_plus/i/group_pickup_4dr/693596}
\item \url{https://www.amazon.sg/Nicokee-Mousepad-Special-Computer-Non-Slip/dp/B08G1GPPWK}
\item \url{https://news.mydrivers.com/1/907/907376.htm}
\item \url{https://read01.com/7RxdyRa.html}
\item \url{https://www.istockphoto.com/vector/funny-red-truck-gm945037348-258131258}
\item \url{https://uk.pinterest.com/rosaleung716/drawing/}
\item \url{https://www.k2r2.com/touxiang_v33935/}
\item \url{http://auction.zhuokearts.com/art/27808163.shtml}
\end{itemize}

\end{document}